\documentclass[10pt,journal,compsoc]{IEEEtran}

\ifCLASSOPTIONcompsoc
  \usepackage[nocompress]{cite}
\else
  \usepackage{cite}
\fi

\usepackage[table,xcdraw]{xcolor}
\usepackage{tabularx}
\usepackage{graphicx}

\usepackage{tikz}
\usepackage{comment}
\usepackage{amsmath,amssymb} 
\usepackage{arydshln}
\usepackage{booktabs}
\usepackage{multirow}
\usepackage[colorlinks,linkcolor=red]{hyperref}
\usepackage{algorithm}  
\usepackage{algorithmicx}  
\usepackage{algpseudocode}  
\usepackage{amsmath}  
\usepackage{array}  
\usepackage{graphicx}
\usepackage{float}
\usepackage{subfig}
 
\usepackage{url}

\begin{document}

\title{Towards Class-wise Fair Adversarial Training via Anti-Bias Soft Label Distillation}

\author{Shiji~Zhao,
        Chi Chen,  Ranjie Duan, Xizhe~Wang, 
        and~Xingxing~Wei$^*$~\IEEEmembership{Member,~IEEE}
\IEEEcompsocitemizethanks{\IEEEcompsocthanksitem Shiji Zhao,  Xizhe Wang, Xingxing Wei were at the Institute of Artificial Intelligence, Beihang University, No.37, Xueyuan Road, Haidian District, Beijing,
100191, P.R. China. (E-mail: \{zhaoshiji123,  xizhewang, xxwei\}@buaa.edu.cn) \hfil\break
\IEEEcompsocthanksitem Chi Chen was at the School of Software, Beihang University, No.37, Xueyuan Road, Haidian District, Beijing,
100191, P.R. China. (E-mail: chenchi@buaa.edu.cn)  \hfil\break
\IEEEcompsocthanksitem Ranjie Duan was with the Security Department, Alibaba Group, Hangzhou 310056, China. (E-mail: ranjieduan@gmail.com)
 \hfil\break
\IEEEcompsocthanksitem Xingxing Wei is the corresponding author.}
}

\markboth{}%
{Shell \MakeLowercase{\textit{et al.}}: Bare Demo of IEEEtran.cls for Computer Society Journals}

\IEEEtitleabstractindextext{
\begin{abstract}
Adversarial Training (AT) is widely recognized as an effective approach to enhance the adversarial robustness of Deep Neural Networks. As a variant of AT, Adversarial Robustness Distillation (ARD) has shown outstanding performance in enhancing the robustness of small models. However, both AT and ARD face robust fairness issue: these models tend to display strong adversarial robustness against some classes (easy classes) while demonstrating weak adversarial robustness against others (hard classes). This paper explores the underlying factors of this problem and points out the smoothness degree of soft labels for different classes significantly impacts the robust fairness from both empirical observation and theoretical analysis. Based on the above exploration, we propose Anti-Bias Soft Label Distillation (ABSLD) within the Knowledge Distillation framework to enhance the adversarial robust fairness. Specifically, ABSLD adaptively reduces the student's error risk gap between different classes, which is accomplished by adjusting the class-wise smoothness degree of teacher's soft labels during the training process, and the adjustment is managed by assigning varying temperatures to different classes. Additionally, as a label-based approach, ABSLD is highly adaptable and can be integrated with the sample-based methods. Extensive experiments demonstrate ABSLD outperforms state-of-the-art methods on the comprehensive performance of robustness and fairness.
\end{abstract}

\begin{IEEEkeywords}
DNNs, Adversarial Training, Knowledge Distillation, Adversarial Robust Fairness.
\end{IEEEkeywords}}

\maketitle

\IEEEdisplaynontitleabstractindextext

%
\IEEEpeerreviewmaketitle

\IEEEraisesectionheading{\section{Introduction}\label{sec:intro}}

\begin{figure*}[t]
  \centering
\includegraphics[width=0.98\textwidth]{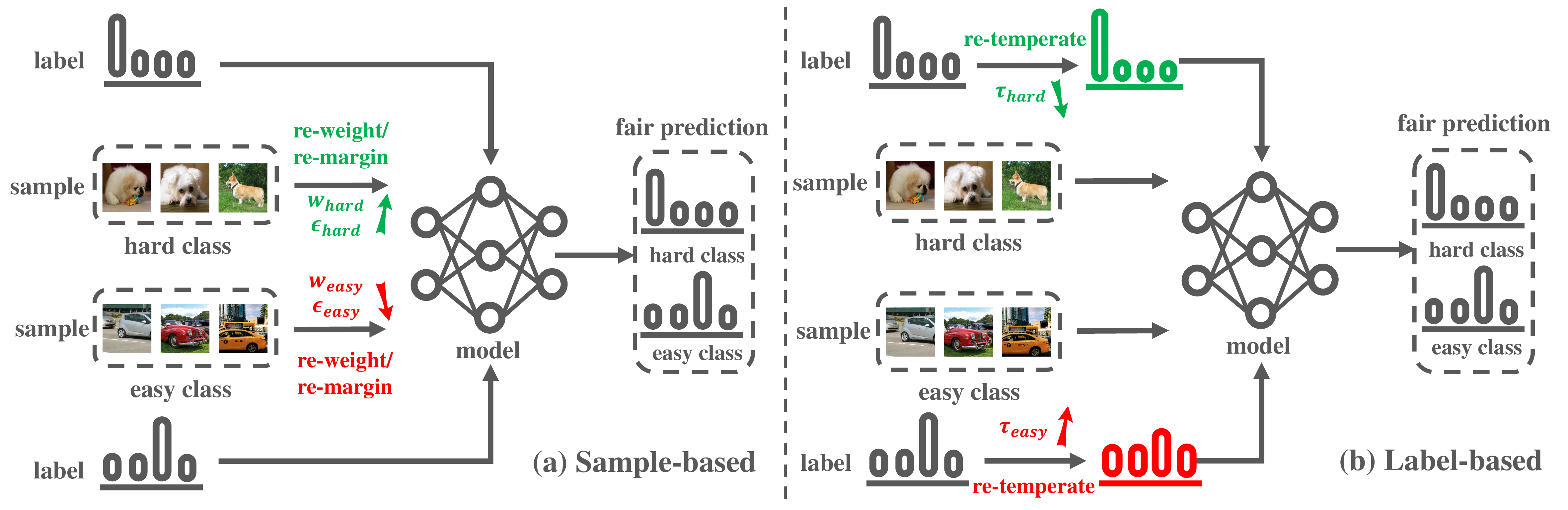}\\
\caption{The comparison between the sample-based fair adversarial training and our label-based fair adversarial training.  For the former ideology in (a), the trained model's bias is avoided by re-weighting the sample's importance or re-margining the sample's adversarial scales for different classes. For the latter ideology in (b), the trained model's bias is avoided by re-temperating the smoothness degree of soft labels for different classes.}
\label{distribution}
\end{figure*}
\IEEEPARstart{D}{eep} neural networks (DNNs) have achieved impressive results in a wide range of tasks, such as classification \cite{he2016deep}, detection \cite{girshick2015fast}, and segmentation \cite{ronneberger2015u}. Despite their success, DNNs are susceptible to adversarial attacks \cite{szegedy2013intriguing,xingxing2023efficient,wei2022adversarial,wei2022simultaneously}, where small perturbations to input data can cause the models to misclassify. To improve the robustness of DNNs against such attacks, Adversarial Training (AT) \cite{madry2017towards,zhang2019theoretically, wang2019improving,jia2023improving} has been introduced and widely validated as an effective defense mechanism against adversarial examples.
To further enhance robustness, Adversarial Robustness Distillation (ARD) \cite{goldblum2020adversarially}, a variant of Adversarial Training (AT), is introduced. ARD leverages Knowledge Distillation (KD) to transfer the robustness of larger models to smaller ones. Subsequent studies \cite{zhu2021reliable,zi2021revisiting,zhao2022enhanced,huang2023boosting,zhao2023mitigating} have demonstrated the outstanding performance of ARD.

While AT and ARD significantly enhance adversarial robustness, studies \cite{benz2021robustness,tian2021analysis,xu2021robust,ma2022tradeoff,sun2023improving,Yue2023Revisiting} have identified a robust fairness issue: these models tend to exhibit strong robustness for certain classes (easy classes) but remain highly vulnerable to others (hard classes). This phenomenon highlights the importance of class-wise security. Although a generally robust model may seem secure for users, poor robust fairness allows attackers to exploit the model’s weaknesses in vulnerable classes, posing serious security threats to real-world applications. In contrast to simply boosting overall robustness, some approaches have been developed to tackle the robust fairness issue in AT and ARD \cite{xu2021robust,wei2023cfa,Yue2023Revisiting,ma2022tradeoff,sun2023improving}, aiming to improve the robustness of the most vulnerable classes without significantly compromising overall robustness. However, the robust fairness problem persists and needs to be further explored.

For that, we thoroughly analyze the factors that influence robust fairness in the optimization objective function. From the perspective of the training sample, the sample introduces a certain degree of biased behavior, mainly reflected in the different learning difficulties and various vulnerabilities to adversarial attacks. For this reason, previous works apply the sample-based ideology (e.g., re-weighting and re-margining) to achieve robust fairness across different classes in the optimization process \cite{xu2021robust, wei2023cfa, Yue2023Revisiting}. 
However, the role of labels, another crucial component in the optimization objective function that guides the model’s training, is often overlooked. Labels can be categorized into one-hot and soft labels, with soft labels extensively studied \cite{szegedy2016rethinking,hinton2015distilling} and proven effective in enhancing the performance of DNNs. Motivated by this, we explore robust fairness from the perspective of soft labels of samples. Notably, we discover that the degree of smoothness in soft labels across different classes (e.g., hard and easy classes) significantly impacts the robust fairness of DNNs, as shown through both empirical observations and theoretical analysis. Intuitively, sharper soft labels indicate stronger supervision, whereas smoother soft labels suggest weaker supervision. Thus, assigning sharper soft labels to hard classes and smoother soft labels to easy classes can be beneficial in enhancing robust fairness.

Based on the above insight, we propose the Anti-Bias Soft Label Distillation (ABSLD) method to address the adversarial robust fairness problem within the knowledge distillation framework. ABSLD adaptively adjusts the smoothness of soft labels by re-tempering the teacher’s soft labels for each class, assigning specific temperatures based on the student’s error risk for that class. For example, if the student shows higher error risk in certain classes, ABSLD assigns lower temperatures to generate sharper soft labels, thereby increasing the student’s learning intensity for those classes compared to others. As a result, the robust error risk gap between different classes is minimized after optimization. Meanwhile, we also explore the combination between sample-based and label-based adversarial training, and we find that ABLSD can be further improved by re-weighting and re-margining methods, which demonstrates ABSLD is highly adaptable. The code can be found in \url{https://github.com/zhaoshiji123/ABSLD}.

Our contribution can be summarized as follows:
\begin{itemize}
\item We investigate the impact of labels on the adversarial robust fairness of DNNs, distinguishing from existing research that primarily focuses on sample perspectives. To the best of our knowledge, we are the first one to find that the smoothness degree of samples' soft labels across different classes can affect robust fairness through both empirical observation and theoretical analysis.
\item We introduce a label-based approach called Anti-Bias Soft Label Distillation (ABSLD) to improve adversarial robust fairness within the knowledge distillation framework. Our method specifically re-temperate the teacher’s soft labels to adjust the smoothness degree for each class, effectively minimizing the error risk gap between classes for the student model. 
\item We combine the sample-based and label-based fair adversarial training into an unifed framework. Experiment show that our ABSLD has the good adaptability and can cooperate with sample-based methods to jointly improve the performance. 
\item We empirically validate the effectiveness of ABSLD through extensive experiments conducted on various datasets and models. Our results show that ABSLD surpasses state-of-the-art methods in terms of robustness and fairness when evaluated against a range of attacks, as measured by the comprehensive metric of Normalized Standard Deviation. 
\end{itemize}

This journal paper is an extended version of our NeurIPS paper \cite{zhao2023improving1}. Compared with the conference version, we have made notable improvement and extension in this version from the following aspects: \textcolor{red}{(1)} At the idea level, we further explore the relationship between  sample-based and label-based fair adversarial training and find they can combine with each other (Section \ref{sec:intro}). \textcolor{red}{(2)} At the method level, we combine an additional existing re-weighting and re-margining method with ABSLD. Meanwhile, we further discuss the inner role of re-temperating method in improving robust fairness (Section \ref{sec:Methods}). \textcolor{red}{(3)} At the experimental level, we give a detailed comparison of the combination version to show the adaptability of ABSLD, and also conduct more experiments to compare with more advanced methods. In addition, we give more detailed ablation studies to comprehensively verify the effectiveness of ABSLD (Section \ref{sec:Experiments}).

The rest of the paper is organized as follows: Related work is given in Section \ref{sec:Related}. Section \ref{sec:Bias} explores the relationship between the class-wise smoothness degree of soft labels and robust fairness. Section \ref{sec:Methods} introduces the details of our ABSLD. The experiments are conducted in Section \ref{sec:Experiments}, and the conclusion is given in Section \ref{sec:Conclusion}.

\section{Related Work}
\label{sec:Related}

\subsection{Adversarial Training}
To protect against adversarial examples, Adversarial Training (AT) \cite{madry2017towards,zhang2019theoretically,wang2019improving,jia2022prior,ruan2023improving,wei2024revisiting} is recognized as an effective approach for obtaining robust models. AT can be expressed as a min-max optimization problem as follows:
\begin{align}
\label{eq:1}
\min\limits_{\theta}E_{(x,y)\sim \mathcal{D}}[\max\limits_{\delta\in\Omega}\mathcal{L} (f(x+\delta;\theta),y)],
\end{align}
where $f(\cdot;\theta)$ denotes a deep neural network characterized by weight $\theta$, while $\mathcal{D}$ signifies a data distribution comprising clean examples $x$ and their corresponding ground truth labels $y$. The term $\mathcal{L}$ refers to the optimization loss function, such as cross-entropy loss. Additionally, $\delta$ represents the adversarial perturbation, and $\Omega$ defines a constraint, which can be expressed as $\Omega = \left\{\delta: ||\delta||\leq \epsilon \right\}$, with $\epsilon$ indicating the maximum scale of perturbation.
To further enhance performance, several variant methods of Adversarial Training (AT) have emerged, including regularization techniques \cite{pang2020boosting,zhang2019theoretically,wang2019improving,jia2024improving}, the use of additional data \cite{sehwag2021robust,rebuffi2021data}, and optimization of the iterative process \cite{jia2022adversarial,rice2020overfitting}. Unlike these approaches that aim to improve overall robustness, this paper focuses on addressing the robust fairness issue.

\subsection{Adversarial Robustness Distillation}

Knowledge Distillation \cite{hinton2015distilling} serves as a training approach that effectively transfers knowledge from large models to smaller ones, and it has found widespread application across various fields. To improve the adversarial robustness of smaller DNNs, Goldblum et al. \cite{goldblum2020adversarially} were the first to introduce Adversarial Robustness Distillation (ARD), which utilizes the clean prediction distribution from robust teacher models to guide the adversarial training of student models. Zhu et al. \cite{zhu2021reliable} contend that the predictions made by the teacher model may lack reliability and propose combining unreliable teacher guidance with student introspection throughout the training process. RSLAD \cite{zi2021revisiting} leverages the teacher's clean prediction distribution to instruct both clean and adversarial examples during training. Meanwhile, MTARD \cite{zhao2022enhanced,zhao2023mitigating} employs both clean and adversarial teachers to simultaneously enhance accuracy and robustness. AdaAD \cite{huang2023boosting} adapts by directly utilizing the teacher's adversarial prediction distribution in the inner maximization to identify optimal match points.
In this paper, we explore how to enhance robust fairness within the framework of knowledge distillation.

\subsection{Adversarial Robust Fairness}
Several researchers have approached the robust fairness issue from various perspectives \cite{xu2021robust,li2023wat,wei2023cfa,Yue2023Revisiting,ma2022tradeoff,wu2021understanding,sun2023improving}, trying to enhance fairness without significantly sacrificing robustness. The most straightforward strategies involve assigning different weights or different adversarial margins to various classes during the optimization process, which adjusts from the sample-based level. 

Specifically, Xu et al. \cite{xu2021robust} introduce Fair Robust Learning (FRL), which modifies loss weights and adversarial margins according to the prediction accuracy of different classes. Ma et al. \cite{ma2022tradeoff} highlight the trade-off between robustness and fairness, proposing Fairly Adversarial Training to address this issue by incorporating a regularization loss that controls the variance of class-wise adversarial error risk. Sun et al. \cite{sun2023improving} present Balance Adversarial Training (BAT) to achieve both source-class fairness (recognizing the differing challenges in generating adversarial examples across classes) and target-class fairness (addressing the unequal tendencies of target classes when generating adversarial examples). Wu et al. \cite{wu2021understanding} suggest that maximum entropy regularization for the model’s prediction distribution can facilitate robust fairness. Wei et al. \cite{wei2023cfa} propose Class-wise Calibrated Fair Adversarial Training (CFA), which tackles fairness by dynamically adjusting adversarial configurations for each class and altering the weight averaging process. To enhance the robust fairness of ARD, Yue et al. \cite{Yue2023Revisiting} introduce Fair-ARD by re-weighting different classes based on their vulnerability levels in the optimization process.  WAT \cite{li2023wat} proposes worst-class adversarial training that leverages no-regret dynamics to enhance robust fairness. DAFA \cite{lee2024dafa} assigns different loss weights and adversarial margins to each class, and tunes them to encourage robustness trade-offs between similar classes. 
In contrast to the sample-based perspective, we aim to address the problem from the label-based perspective.

\section{Robust Fairness via Smoothness Degree of Soft Labels}
\label{sec:Bias}

As a component of model optimization, the label information used to guide the model plays an important role. Labels can be categorized into one-hot labels, which convey information about only one class, and soft labels, which are recognized as an effective means to reduce overfitting and enhance performance \cite{szegedy2016rethinking}.
Many previous approaches tend to directly apply soft labels but neglect a more in-depth analysis towards the class-wise smoothness of soft labels, either by employing a uniform smoothness degree \cite{szegedy2016rethinking} or by indiscriminately adjusting the smoothness degree of soft labels across all classes \cite{hinton2015distilling} without careful consideration in class-wise level. Different from these previous methods, we investigate the following question: \emph{if we adjust the class-wise smoothness degree of soft labels, will it influence the class-wise robust fairness of the trained model?} Intuitively, varying the smoothness of soft labels implies different levels of supervision intensity, suggesting that fairness may be achievable through such adjustments. Here we try to explore the relationship between the class-wise smoothness degree of soft labels and robust fairness through both empirical observation and theoretical analysis.

\subsection{Empirical Observation}
\begin{figure}
\centering
\subfloat[Class-wise and average Robustness of ResNet-18.]{\label{figure_ca}\includegraphics[width=0.48\linewidth]{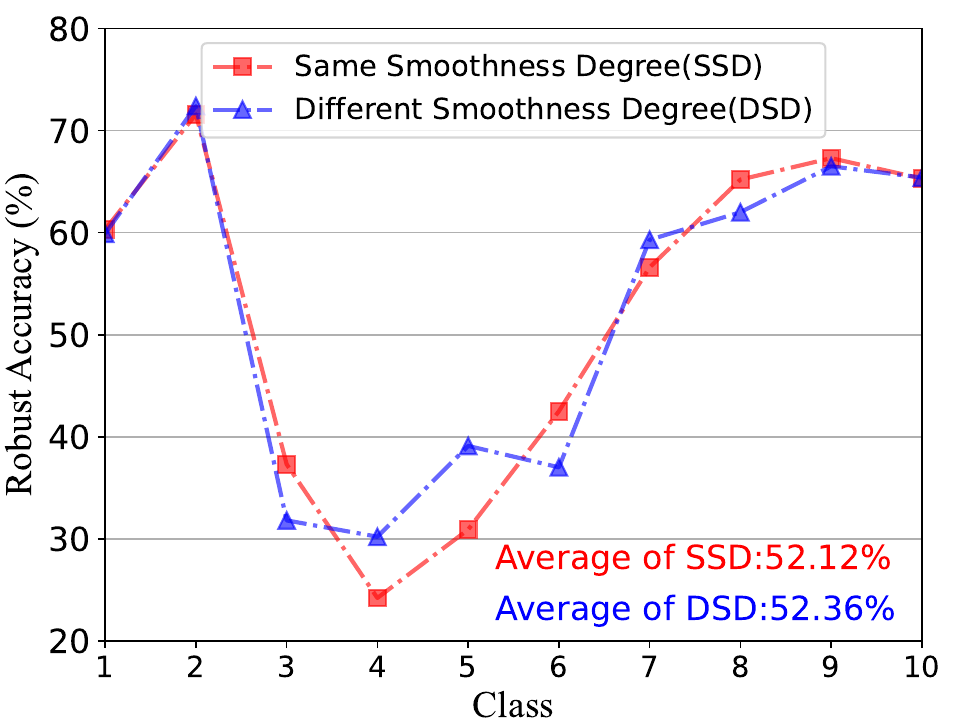}}
\subfloat[Class-wise and average Robustness of MobileNet-v2.]{\label{figure_cb}\includegraphics[width=0.48\linewidth]{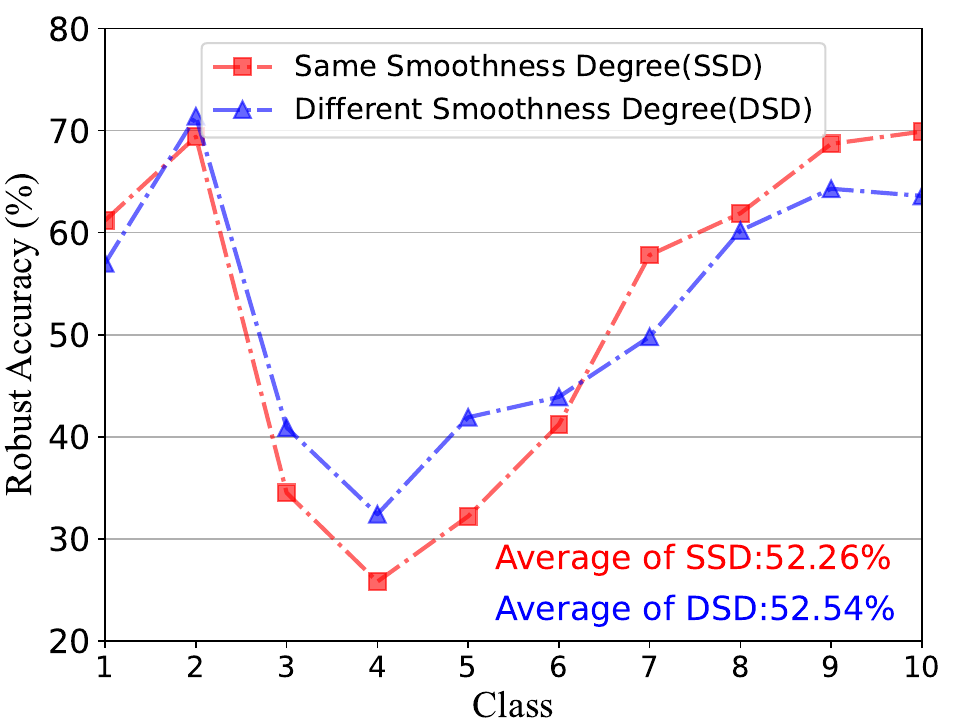}}
\caption{The class-wise and average robustness of DNNs guided by soft labels with the same smoothness degree (SSD) and different smoothness degree (DSD) for different classes, respectively. Specifically, sharper soft labels are applied to hard classes, while smoother soft labels are used for easy classes. Here we train two DNN architectures, ResNet-18 and MobileNet-v2 using SAT \cite{madry2017towards} on the CIFAR-10 dataset, with robust accuracy evaluated using PGD. The model checkpoint is selected based on the highest mean value of both the overall average robustness and the worst class robustness, following \cite{wei2023cfa}. The results show that while the blue and red lines exhibit similar average robustness, the worst class robustness of the blue lines is significantly improved compared to the red lines. }
\label{fig:result}
\end{figure}

Here, we examine how the class-wise smoothness of soft labels influences adversarial training. We begin by training the model with soft labels that have a consistent smoothness degree across all classes (using a smoothing coefficient\footnote{\scriptsize For N-class one-hot ground truth labels, applying a smoothing coefficient $\gamma$ reduces the highest probability (correct class) from 1 to $1-\gamma$, while the probabilities of incorrect classes increase from 0 to $\frac{\gamma}{n-1}$.} of 0.2). Subsequently, we assign distinct smoothness degrees to soft labels for hard and easy classes: sharper soft labels (smoothing coefficient of 0.05) are used for hard classes, while smoother soft labels (smoothing coefficient of 0.35) are applied to easy classes. The experiment is conducted using SAT \cite{madry2017towards}, as shown in Figure \ref{fig:result}. 

The results demonstrate that the class-wise smoothness degree of soft labels significantly influences robust fairness across classes. Specifically, applying sharper soft labels to hard classes and smoother soft labels to easy classes helps alleviate the class-wise robust fairness issue. For instance, in ResNet-18, the robust accuracy for the two most challenging classes (classes 4 and 5) improves from 24.2\% and 30.9\% with the same smoothness degree to 30.2\% and 39.1\% with different smoothness degrees. This adjustment leads to a noticeable enhancement in class-wise robust fairness, while the average robust accuracy sees a slight improvement (52.12\% vs. 52.36\%). Similar results are observed in MobileNet-v2, suggesting that appropriately adjusting the smoothness degree of soft labels for each class can effectively enhance robust fairness.

\subsection{Theoretical Analysis}
Here we aim to theoretically analyze how the smoothness degree of soft labels affects class-wise fairness. We start by examining the model's bias performance when guided by soft labels with a uniform smoothness degree. Following this, we provide Corollary \ref{Assumption}, which extends the prediction distribution of a binary linear classifier to the prediction distribution of DNNs according to the theoretical analysis from \cite{xu2021robust} and \cite{ma2022tradeoff}.

\newtheorem{thm3}{Corollary}
\begin{thm3}\label{corollary1}
\label{Assumption}
\emph{A dataset $(x,y)\sim\mathcal{D}$ contains $2$ classes (hard class $c_+$ and easy class $c_-$). Based on the label distribution $y$, the soft label distribution with same smoothness degree $P_{\lambda1}=\{p_{c_+}^{\lambda1},p_{c_-}^{\lambda1}\}$ can be generated and satisfies:
\begin{align}
\label{sec3:eq-1}
1>p_{c_-}^{\lambda1}(x_{c_-}) = p_{c_+}^{\lambda1}(x_{c_+}) >0.5,  
\end{align}
If a DNN model $f$ is optimized by minimizing the average optimization error risk in $\mathcal{D}$ with the guidance of the equal soft labels $P_{\lambda1}=\{p_{c_+}^{\lambda1},p_{c_-}^{\lambda1}\}$, and obtain the relevant parameter $\theta_{\lambda1}$, where the optimization error risk is measured by Kullback–Leibler divergence loss ($KL$):
\begin{align}
\label{sec3:eq-2}
f(x;\theta_{\lambda1})=\mathop{\arg\min}_{f}\mathbb{E}_{(x,y)\sim \mathcal{D}}( KL(f(x;\theta_{\lambda1}); P_{\lambda1})),
\end{align}
then the error risks (the expectation that samples are wrongly predicted by the model) for classes $c_+$ and $c_{-}$  have a relationship as follows:
\begin{align}
\label{sec3:eq-3}
 R({f(x_{c_+};\theta_{\lambda1})}) >  R({f(x_{c_-};\theta_{\lambda1})}),
\end{align}
where the error risks can be defined:
\begin{align}
\label{sec3:eq-4}
R({f(x_{c_+};\theta_{\lambda1})}) = \mathbb{E}_{(x,y)\sim \mathcal{D}}(CE(f(x_{c_{+}};\theta_{\lambda1}); y_{c_+})),\notag \\
R({f(x_{c_-};\theta_{\lambda1})}) = \mathbb{E}_{(x,y)\sim \mathcal{D}}(CE(f(x_{c_{-}};\theta_{\lambda1}); y_{c_-})).
\end{align}
}
\end{thm3}

Corollary \ref{Assumption} shows that optimizing hard and easy classes with equal intensity will inevitably lead to model bias, which primarily arises from the intrinsic characteristics of the samples rather than the optimization approach. Based on Corollary \ref{Assumption}, we further examine the performance differences influenced by various types of soft labels. Theorem \ref{thm1} is presented to elucidate the relationship between the class-wise smoothness degree of soft labels and fairness.

\newtheorem{thm}{Theorem}
\begin{thm}\label{thm1}
\emph{Following the setting in Corollary \ref{Assumption}, for a dataset $\mathcal{D}$ containing $2$ classes ($c_+$ and $c_-$), two soft label distribution ($P_{\lambda1}=\{p_{c_+}^{\lambda1},p_{c_-}^{\lambda1}\}$ and $P_{\lambda2}=\{p_{c_+}^{\lambda2},p_{c_-}^{\lambda2}\}$) exist, where $P_{\lambda2}$ has a correct prediction distribution but has a limited different class-wise smoothness degree of soft labels ($v_1>0$, $v_2>0$):
\begin{align}
\label{sec3:eq-5}
1>&p_{c_+}^{\lambda2}(x_{c_+}) =p_{c_+}^{\lambda1}(x_{c_+}) + v_1 >p_{c_+}^{\lambda1}(x_{c_+})=  \notag \\ &p_{c_-}^{\lambda1}(x_{c_-})  > p_{c_-}^{\lambda2}(x_{c_-})= p_{c_-}^{\lambda1}(x_{c_-}) - v_2 >0.5, 
\end{align}
then the model is trained with the guidance of the soft label distribution $P_{\lambda1}$ and soft label distribution  $P_{\lambda2}$ and obtains the trained model parameters $\theta_{\lambda1}$ and $\theta_{\lambda2}$, respectively. If the model parameter $\theta_{\lambda2}$ still satisfies: $ R({f(x_{c_+};\theta_{\lambda2})}) >  R({f(x_{c_-};\theta_{\lambda2})})$, then the model's error risk for hard classes $c_+$ and easy classes $c_{-}$  has a relationship as follows:
\begin{align}
\label{sec3:eq-6}
 R(f(x_{c_+};\theta_{\lambda1}))-&R(f(x_{c_-};\theta_{\lambda1})) > \notag \\ 
 &R(f(x_{c_+};\theta_{\lambda2}))-R(f(x_{c_-};\theta_{\lambda2})).
\end{align}
}
\end{thm}

The proof of Theorem \ref{thm1} can be found in Appendix A. Based on Theorem \ref{thm1}, the class-wise smoothness degree of soft labels theoretically has an impact on class-wise robust fairness. If the soft label distribution with different smoothness degree $P_{\lambda2}$ is applied to guide the model training, where the sharper smoothness degree of soft labels for hard classes and smoother smoothness degree of soft labels for easy classes, the model will appear smaller error risk gap between easy and hard class compared with the soft label distribution with same smoothness degree $P_{\lambda1}$, which demonstrates better robust fairness. The Theorem \ref{thm1} theoretically demonstrates if we appropriately adjust the class-wise smoothness degree of soft labels, the model can achieve class-wise robust fairness.

\section{Anti-Bias Soft Label Distillation}
\label{sec:Methods}
\subsection{Overall Framework}
Based on the above finding, adjusting the class-wise smoothness degree of soft labels can be seen as a potential approach to achieve robust fairness. To leverage this concept, we consider integrating it into Adversarial Robustness Distillation (ARD), a well-established method for enhancing the robustness of smaller models \cite{zhu2021reliable,zi2021revisiting,zhao2022enhanced,huang2023boosting,zhao2023mitigating}. ARD's core principle involves guiding the student model's optimization process using the teacher's soft labels, which is highly consistent with our idea of improving model fairness by adjusting the smoothness degree of soft labels. Therefore, further improvements based on ARD could obtain a student model that achieves both strong robustness and fairness.

Here we introduce the Anti-Bias Soft Label Distillation (ABSLD) method to achieve a student model with adversarial robust fairness. The student model's bias can be assessed by the optimization error risk gap between different classes. The optimization objective function for ABSLD can be formulated as follows:
\begin{align}
\label{sec4:eq-1}
    &\mathop{\arg\min}_{f_s}\mathbb{E}_{(x,y)\sim \mathcal{D}}(\mathcal{L}_{absld}(\Tilde{x},x;f_s,f_t^{'})), \\
&s.t. ~R(f_s(\Tilde{x}_k))=\frac{1}{C}\sum_{i=1}^CR(f_s(\Tilde{x}_i)), 
\end{align}
where $x_i$ and $\Tilde{x}_i$ represent the clean and adversarial examples for the $i$-th class, respectively. $f_s$ refers to the student model, while $f_t^{'}$ denotes the teacher model utilizing Anti-Bias Soft Labels, $C$ represents the total number of classes, $\mathcal{L}_{absld}$ is the loss function, and $R(f_s(\Tilde{x}_k))$ denotes the robust error risk for the $k$-th class in student model $f_s$. To measure the robust error risk $R(f_s(\Tilde{x}_k))$, we use the Cross-Entropy loss following \cite{ma2022tradeoff}. 

\subsection{Re-temperate Teacher's Soft Labels}
To modify the class-wise smoothness degree of soft labels in ARD, there are two possible approaches: one involves utilizing student feedback to update the teacher parameters while optimizing the students. However, this ideology is supposed to retrain the teacher model, which can lead to significant optimization challenges and increased computational costs. The alternative approach is to directly adjust the smoothness degree of soft labels across different classes. Inspired by \cite{zhao2023mitigating}, we utilize temperature as a tool to directly manage the smoothness degree of soft labels throughout the training process. Here we present Theorem \ref{thm2}, which discusses the relationship between the teacher's temperature and the class-wise error risk gap of the student.

\newtheorem{thm2}{\bf Theorem}[section]
\begin{thm}\label{thm2}
\emph{If teacher $f_{t}^{'}$ has a correct prediction distribution, teacher temperature $\tau_{c+}^t$ of hard class $c+$ is positively correlated with the error risk gap for student $f_{s}$, and teacher temperature $\tau_{c-}^t$ of easy class $c-$ is negatively correlated with the error risk gap for student $f_{s}$.}
\end{thm} 

The proof for Theorem \ref{thm2} is provided in Appendix B. Specifically, in line with the findings in \cite{Yue2023Revisiting}: The teacher has a more correct prediction distribution than the student even in the worst classes, which indicates that Theorem \ref{thm2} is valid in the majority of scenarios. Theorem \ref{thm2} illustrates how varying temperatures can affect the robust fairness of the student model: when the error risk for the $k$-th class exceeds the average error risk, we interpret this class as relatively difficult compared to others. Consequently, the teacher's temperature for the $k$-th class will decrease, resulting in sharper smoothness for the soft labels. This adjustment will subsequently increase the optimization gap between the teacher's distribution and the student's distribution for the $k$-th class, leading to a stronger learning intensity for that class and ultimately reducing the class-wise optimization error risk gap of the student.

To attain the optimization objective mentioned above, we modify the teacher's temperature for the $k$-th class, denoted as $\Tilde{\tau}_{k}^t$ to guide the adversarial examples as follows:
\begin{align}
\label{sec4:eq-2}
\Tilde{\tau}_{k}^t=\Tilde{\tau}_{k}^t - \eta_{\tau} \cdot \frac{R(f_{s}(\Tilde{x}_k))-\frac{1}{C}\sum_{i=1}^CR(f_{s}(\Tilde{x}_i))}{max(|R(f_{s}(\Tilde{x}_k))-\frac{1}{C}\sum_{i=1}^CR(f_{s}(\Tilde{x}_i))|)}, 
\end{align}
where the parameter $\eta_{\tau}$ denotes the learning rate, while $max(.)$ refers to the operation of taking the maximum value, and $|.|$ signifies the absolute value operation. The expression $max(|.|)$ is used for regularization purposes to ensure the stability of the optimization process. The update operation described in Eq.(\ref{sec4:eq-2}) allows the teacher temperature $\Tilde{\tau}_{k}^t$ to adjust according to the difference between the error risk $R(f_{s}(\Tilde{x}_k))$ of the $k$-th class for the student model, and the average error risk $\frac{1}{C}\sum_{i=1}^C R(f_{s}(\Tilde{x}_i))$.

As claimed in \cite{xu2021robust}, fairness issue exists in both clean and adversarial examples, and these problems can influence one another, making it essential to ensure fairness for both types of data. Given that clean and adversarial examples from the same class can exhibit different error risks during training, using the same set of class temperatures for those two examples is not appropriate. To address this, we optimize the student's clean error risk $R(f_s(x_k))$ and the robust error risk $R(f_s(\Tilde{x}_k))$ simultaneously. Specifically, we employ two distinct sets of teacher temperatures: $\tau_{k}^{t}$ for clean examples and $\Tilde{\tau}_{k}^{t}$ for adversarial examples, to adjust the teacher's soft labels, respectively.

\begin{algorithm}[t]  
  \caption{Overview of ABSLD}  
  \label{algorithm:1}
  \begin{algorithmic}[1]   
   \Require {the trainset $\mathcal{D}$, Student $f_s$ with random initial weight $\theta_{s}$ and temperature $\tau_{s}$, pretrained robust teacher $f_t$, the initial temperature $\tau_{y}^{t}$ and $\Tilde{\tau}_{y}^{t}$ for the teacher's soft labels for clean examples $x$ and adversarial examples $\Tilde{x}$, where $y=\{1,\dots,C\}$, the max training epochs $max$-$epoch$.}
     \For{$0$ to $max$-$epoch$} 
        \For{$k ~ ~in ~ ~y=\{1,\dots,C\}$}  
       \State { \small $R(f_s(\Tilde{x}_k))=0,~R(f_s(x_k)) =0$.}\emph{\small~ //~ Initialize the clean and robust error risk for each class.} 
        \EndFor \For{$Every~minibatch(x,y)~in~\mathcal{D}$}  
            \State { \small $\Tilde{x}=\mathop{argmax}\limits_{||\Tilde{x}-x|| \leq \epsilon} KL(f_s(\Tilde{x};\tau^{s}),f_{t}^{'}(x;\Tilde{\tau}_{y}^{t}))$.} \emph{\small~ //~ Get adversarial examples with teacher's soft labels.} 
          \State { \small $\theta_{s} = \theta_{s} - \eta \cdot \nabla_{\theta}\mathcal{L}_{absld}(\Tilde{x},x;f_s,f_{t}^{'})$.} \emph{\small~ //~ Update student weight $\theta_{s}$ with teacher's soft labels.}  
          \State { \small $R(f_s(\Tilde{x}_{y}))=R(f_s(\Tilde{x}_{y})) + CE(f_s(\Tilde{x}),{y})$.}  \emph{\small~ //~ Calculate robust error risk for each class.}
          \State { \small $R(f_s(x_{y}))=R(f_s(x_{y})) + CE(f_s({x}),{y})$.}  \emph{\small~ //~ Calculate clean error risk for each class.}
          \EndFor  
        \For{$k ~ ~in ~ ~y=\{1,\dots,C\}$}  
        \State { \small Update $\Tilde{\tau}_{k}^t$ and ${\tau}_{k}^t$ based on Eq.(\ref{sec4:eq-2}).} \emph{\small~ //~ Adjust teacher's soft labels for $\Tilde{x}$ and $x$.}
        \EndFor 
    \EndFor  
  \end{algorithmic}  
\end{algorithm}  

Then we extend the Anti-Bias Soft Label Distillation based on \cite{zi2021revisiting} and the loss function $\mathcal{L}_{absld}$ in Eq.(\ref{sec4:eq-1}) can be formulated as follows:
\begin{align}
\label{sec4:eq-3}
\mathcal{L}_{absld}&(\Tilde{x},x;f_s,f_{t}^{'}) = \alpha \frac{1}{C} \sum_{i=1}^{C} KL(f_s(\Tilde{x}_i;\tau^s), f_{t}^{'}({x}_i;\Tilde{\tau}_{i}^t))\notag \\ 
 &+ (1-\alpha) \frac{1}{C} \sum_{i=1}^{C}  KL(f_s({x}_i;\tau^s), f_{t}^{'}({x}_i;\tau_{i}^t)),
\end{align}
where $KL$ represents Kullback–Leibler divergence loss, $\alpha$ is the trade-off parameter between accuracy and robustness, $f(x;\tau)$ denotes model $f$ predicts the output probability of $x$ with temperature $\tau$ in the final softmax layer. It should be mentioned that the teacher is frozen and we apply the teacher's predicted soft labels $f_{t}^{'}(x_k;\Tilde{\tau}_k^{t})$ for $k$-th class to generate adversarial examples $\Tilde{x}_k$ as follows:
\begin{align}
\label{sec4:eq-4}
\Tilde{x}_k=\mathop{argmax}\limits_{||\Tilde{x}_k-x_k|| \leq \epsilon} KL(f_s(\Tilde{x}_k;\tau^{s}),f_{t}^{'}(x_k;\Tilde{\tau}_k^{t})),
\end{align}
and the complete process can be viewed in Algorithm \ref{algorithm:1}.

In addition, to maintain stability when adjusting the teacher prediction distribution, we hold the student temperature $\tau^s$ constant, and the student's optimization error risk for each class can be compared under the same standard.

\subsection{Combining Sample-based with Label-based Ideas}

In addition, due to the different implementation paths, the sample-based and label-based ideas intuitively do not conflict with each other. Here we try to confirm our re-temperating method can combine with the re-weighting and re-margining method. 

As for the combination with the re-weighting method, we increase the optimization weights for hard classes and decrease the optimization weights for easy classes. To keep the same optimization goal with the re-temperating, we adjust the $k$-th class weight $\Tilde{w}_{k}$ in each epoch as follows:
\begin{align}
\label{sec4:eq-5}
\Tilde{w}_{k}=\Tilde{w}_{k} + \eta_{w} \cdot \frac{R(f_{s}(\Tilde{x}_k))-\frac{1}{C}\sum_{i=1}^CR(f_{s}(\Tilde{x}_i))}{max(|R(f_{s}(\Tilde{x}_k))-\frac{1}{C}\sum_{i=1}^CR(f_{s}(\Tilde{x}_i))|)}, 
\end{align}
where $\eta_{w}$ is the learning rate for class-wise weights, $\Tilde{w}_{k}$ is the optimization weights of $k$-th class for adversarial examples. The update operation in Eq.(\ref{sec4:eq-5}) can adaptively change optimization weights. Also, we apply two different sets of optimization weights $w_{k}$ and $\Tilde{w}_{k}$ towards the clean examples and adversarial examples, respectively. And the optimization function $\mathcal{L}_{absld+rw}$ of ABSLD+Re-Weight can be formulated as follows:
\begin{align}
\label{sec4:eq-6}
&\mathcal{L}_{absld+rw}(\Tilde{x},x;f_s,f_{t}^{'}) = \alpha \frac{1}{C} \sum_{i=1}^{C}\Tilde{w}_{k} KL(f_s(\Tilde{x}_i;\tau^s), f_{t}^{'}({x}_i;\Tilde{\tau}_{i}^t))\notag \\ 
 &+ (1-\alpha) \frac{1}{C} \sum_{i=1}^{C} w_{k} KL(f_s({x}_i;\tau^s), f_{t}^{'}({x}_i;\tau_{i}^t)).
\end{align}

As for the combination with the re-margining method, we increase the maximum adversarial perturbation bound for the hard class and slightly reduce the maximum adversarial perturbation bound for the easy class following the previous method \cite{xu2021robust}. Here we re-margin the adversarial scale $\epsilon_{k}$ of $k$-th class in each epoch and can formulate it as follows:
\begin{align}
\label{sec4:eq-7}
\Delta_{k}^{bndy} =&\Delta_{k}^{bndy} +\eta_{\epsilon} \cdot \big(R(f_{s}(\Tilde{x}_k))-R(f_{s}(x_k))-\notag \\ 
&\frac{1}{C}\sum_{i=1}^CR(f_{s}(\Tilde{x}_i))+\frac{1}{C}\sum_{i=1}^CR(f_{s}(x_i))\big), 
\end{align}
\begin{align}
\label{sec4:eq-8}
\epsilon_{k} = \epsilon_{k}\cdot exp\big(\beta\cdot(\Delta_{k}^{bndy}-\gamma)\big), 
\end{align}
where $\Delta_{k}^{bndy}$ denotes the boundary error risk gap \cite{zhang2019theoretically} between $k$-th class and all the class (boundary error risk denotes the gap between clean error risk and the robust error risk), $\eta_{\epsilon}$ is the learning rate for re-margining operation, $\beta$ and $\gamma$ are two constant hyper-parameters.



Here we want to further discuss the role of sample-based methods and label-based methods. As for the sample-based method, the re-weighting can play the role of effectively adjusting the optimization strength for different classes in the training process, while the re-margining can directly adjust the sample difficulty. As for the label-based method, in addition to adjusting the optimization strength at the label level, re-temperating is more like an upgrade of soft label methods, while soft label methods can be applied to reduce the overfitting of unnecessary information by introducing the smoothness degree for one-hot labels. We guess the re-temperating method fully utilizes this type of superiority and reasonably decouples the superiority from the level of the entire data to the level of each class, and further improves the adversarial robust fairness.

\begin{figure}[t]
  \centering
\subfloat[\centering Error risk gap between trainset and testset in the checkpoint of 200-th epoch.]{\label{overfit1}\includegraphics[width=0.48\linewidth]{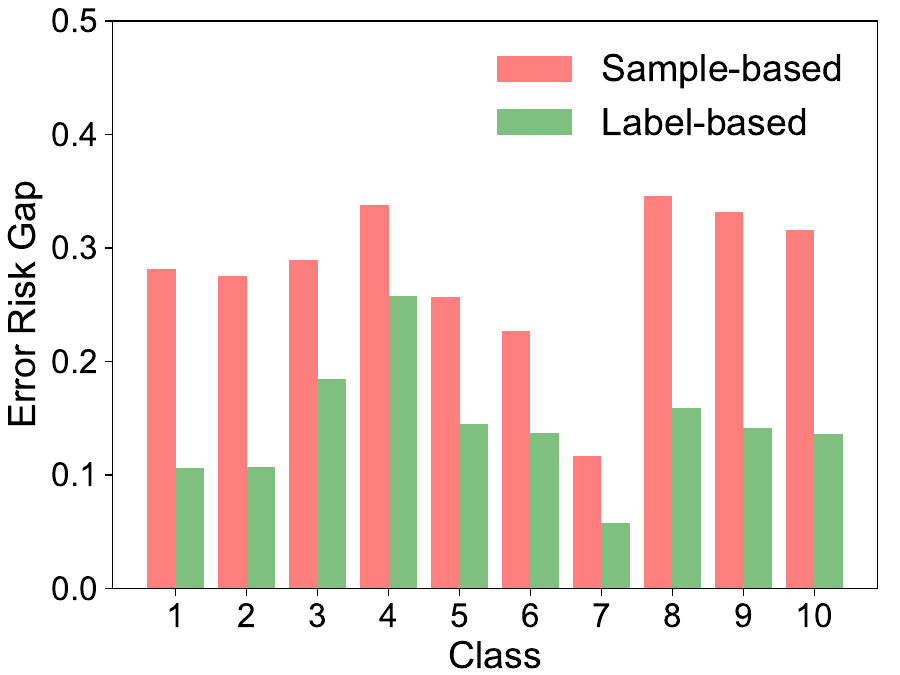}}
\subfloat[\centering Error risk gap between trainset and testset in the checkpoint of 300-th epoch.]{\label{overfit2}\includegraphics[width=0.48\linewidth]{picture/overfitting_200_new.pdf}}
\caption{The error risk gap between trainset and testset between sample-based method and label-based method in different training epochs. The result demonstrates that label-based method remarkably alleviates the overfitting phenomenon during the training process.}
\label{overfit}
\end{figure}

\begin{table*}[ht] \scriptsize
\caption{Result in average robustness(\%) (Avg.$\uparrow$), worst robustness(\%) (Worst$\uparrow$), and normalized standard deviation (NSD$\downarrow$) on CIFAR-10 of ResNet-18.}
\centering  
\scalebox{1.05}{ 
\begin{tabular}{c|ccc|ccc|ccc|ccc|ccc}
\hline
\multirow{2}{*}{Method} & \multicolumn{3}{c|}{Clean}     & \multicolumn{3}{c|}{FGSM}      & \multicolumn{3}{c|}{PGD} & \multicolumn{3}{c|}{CW$_{\scriptscriptstyle \rm{\infty}}$}      & \multicolumn{3}{c}{AA}         \\ \cline{2-16} 
                        & Avg.  & Worst & NSD        & Avg.  & Worst & NSD        & Avg.   & Worst & NSD        & Avg.  & Worst & NSD        & Avg.  & Worst & NSD        \\  \hline 
Natural                 & 94.57 & 86.30 & 0.035 & 18.60 & 9.00  & 0.436          & 0      & 0     & -              & 0     & 0     & -              & 0     & 0     & -              \\ 
SAT\cite{madry2017towards}                     & 84.03 & 63.90 & 0.118          & 56.71 & 26.70 & 0.283          & 49.34  & 21.00 & 0.332          & 48.99 & 19.90 & 0.352          & 46.44 & 16.80 & 0.385          \\
TRADES\cite{zhang2019theoretically}                  & 81.45 & 67.60 & 0.113          & 56.65 & 36.60 & 0.267          & 51.78  & 30.40 & 0.301          & 49.15 & 27.10 & 0.341          & 48.17 & 25.90 & 0.350          \\
RSLAD\cite{zi2021revisiting}                   & 82.94 & 66.30 & 0.122          & 59.51 & 34.70 & 0.244          & 54.00  & 28.50 & 0.276          & \textbf{52.51} & 27.00 & 0.296          & \textbf{51.25} & 25.50 & 0.304          \\
AdaAD\cite{huang2023boosting}                   & 84.73 & 68.10 & 0.114          & 59.70 & 34.80 & 0.246          & 53.82  & 29.30 & 0.285          & 52.30 & 26.00 & 0.312          & 50.91 & 24.70 & 0.322          \\  \hdashline
FRL\cite{xu2021robust}                     & 82.29 & 64.60 & 0.114          & 55.03 & 37.10 & 0.230          & 49.05  & 31.70 & 0.248          & 47.88 & 30.40 & 0.266          & 46.54 & 28.10 & 0.280          \\
BAT\cite{sun2023improving}                     & 86.72 & 72.30 & 0.092          & \textbf{60.97} & 33.80 & 0.255          & 49.60  & 22.70 & 0.325          & 47.49 & 19.50 & 0.354          & 48.18 & 20.70 & 0.341          \\
CFA\cite{wei2023cfa}                     & 78.64 & 63.60 & 0.123          & 57.95 & 36.80 & 0.231          & 54.42  & 33.30 & 0.258          & 50.91 & 27.50 & 0.288          & 50.37 & 26.70 & 0.296          \\
WAT\cite{li2023wat} & 81.07 & 69.40 & 0.078 & 55.45 & 39.60 & 0.215 & 49.10 & 35.80 & 0.232 & 46.92 & 30.90 & 0.268 & 45.97 & 29.50 & 0.276  \\
DAFA\cite{lee2024dafa}                     & 82.12 & 67.60 & 0.096          & 57.15 & 39.80 & 0.213          & 51.94  & 35.60 & 0.242          & 48.99 & 31.10 & 0.285          & 48.05 & 29.00 & 0.297          \\
Fair-ARD\cite{Yue2023Revisiting} & 83.81 & 69.40 & 0.112          & 58.41 & 38.50 & 0.251          & 50.91  & 29.90 & 0.296          & 49.96 & 28.30 & 0.312 & 47.97 & 25.10 & 0.338    \\
\textbf{ABSLD}           & 83.04 & 68.10 & 0.103         & 59.83 & \textbf{40.50} & \textbf{0.202} & \textbf{54.50}  & \textbf{36.50}  & \textbf{0.216} & 51.77 & \textbf{32.80} & \textbf{0.249} & 50.25 & \textbf{31.00} & \textbf{0.256} \\ 
\hline
\end{tabular}
} \label{tab:1}
\end{table*}

\begin{table*}[ht] \scriptsize
\caption{Result in average robustness(\%) (Avg.$\uparrow$), worst robustness(\%) (Worst$\uparrow$), and normalized standard deviation (NSD$\downarrow$) on CIFAR-10 of MobileNet-v2.}
\centering  
\scalebox{1.05}  { 
\begin{tabular}{c|ccc|ccc|ccc|ccc|ccc}
\hline
\multirow{2}{*}{Method} & \multicolumn{3}{c|}{Clean}     & \multicolumn{3}{c|}{FGSM}      & \multicolumn{3}{c|}{PGD} & \multicolumn{3}{c|}{CW$_{\scriptscriptstyle \rm{\infty}}$}      & \multicolumn{3}{c}{AA}         \\ \cline{2-16} 
                        & Avg.  & Worst & NSD        & Avg.  & Worst & NSD        & Avg.   & Worst & NSD        & Avg.  & Worst & NSD        & Avg.  & Worst & NSD        \\ \hline
Natural                 & 93.35 & 85.20 & 0.036 & 12.21 & 0.90  & 0.750           & 0      & 0     & -              & 0     & 0     & -              & 0     & 0     & -              \\
SAT\cite{madry2017towards}                     & 82.30 & 65.80 & 0.131          & 56.19 & 34.60 & 0.279          & 48.52  & 24.90 & 0.334          & 47.22 & 23.80 & 0.361          & 44.38 & 18.50 & 0.410          \\
TRADES\cite{zhang2019theoretically}                   & 79.37 & 59.00 & 0.131          & 54.94 & 31.60 & 0.297          & 50.03  & 27.20 & 0.330          & 47.02 & 23.20 & 0.367          & 46.25 & 22.10 & 0.379          \\
RSLAD\cite{zi2021revisiting}                   & 82.96 & 66.30 & 0.130          & \textbf{59.84} & 34.30 & 0.256          & \textbf{53.88}  & 28.20 & 0.288          & \textbf{52.28} & 25.70 & 0.316          & 50.67 & 23.90 & 0.333          \\
AdaAD\cite{huang2023boosting}                   & 83.72 & 66.90 & 0.123          & 57.63 & 33.70 & 0.262          & 51.89  & 26.90 & 0.300          & 50.34 & 24.70 & 0.317          & 48.81 & 22.50 & 0.334          \\ \hdashline
FRL\cite{xu2021robust}                      & 81.02 & 70.50 & 0.089          & 53.84 & 40.50 & 0.207          & 47.71  & 35.00 & 0.241          & 44.96 & 30.90 & 0.276          & 43.53 & 28.40 & 0.292          \\
BAT\cite{sun2023improving}                      & 83.01 & 70.30 & 0.102          & 53.01 & 32.10 & 0.284          & 44.08  & 25.00 & 0.344          & 41.85 & 21.60 & 0.397          & 42.65 & 23.40 & 0.369          \\
CFA\cite{wei2023cfa}                     & 80.34 & 64.60 & 0.112          & 56.45 & 32.20 & 0.260          & 52.34  & 28.10 & 0.292          & 48.62 & 23.20 & 0.320          & \textbf{50.68} & 23.90 & 0.331          \\ 
WAT\cite{li2023wat} & 78.05 & 66.60 & 0.083 & 51.45 & 39.70 & 0.219 & 47.18 & 34.70 & 0.238 &  43.47 & 28.60 & 0.267 & 42.73 & 27.30 & 0.276 \\
DAFA\cite{lee2024dafa}                     & 79.45 & 64.10 & 0.101          & 53.95 & 38.60 & 0.228          & 49.11  & 34.40 & 0.259          & 46.31 & 29.50 & 0.299          & 45.41 & 28.00 & 0.310          \\
Fair-ARD\cite{Yue2023Revisiting}  & 82.44 & 69.40 & 0.100          & 56.29 & 38.60 & 0.226          & 50.91 & 29.90 &  0.263 & 48.18 & 30.80 & 0.286 & 46.62 & 27.60 & 0.302 \\
\textbf{ABSLD}           & 82.54 & 69.50 & 0.102          & 58.55 & \textbf{41.40} & \textbf{0.207} & 52.99  & \textbf{35.70} & \textbf{0.224} & 50.39 & \textbf{31.90} & \textbf{0.254} & 48.71 & \textbf{30.30} & \textbf{0.259} \\ \hline
\end{tabular}
} \label{tab:2}
\end{table*}

To verify the above opinion, we directly compare the class-wise error risk gap between the trainset and testset for the sample-based method and label-based method. The larger error risk gap denotes the less serious the overfitting. The error risk is obtained based on Cross Entropy Loss. We select several typical checkpoints on CIFAR-10 of ResNet-18 in different training periods (the checkpoint in the 200-th and 300-th training epoch). The results are shown in Figure \ref{overfit}: It can be seen that compared with the sample-based method, the label-based method has a smaller error risk gap between the trainset and testset for all the classes, which shows that the label-based method can effectively suppress overfitting in class-wise level. Therefore, The optimization results of sample-based methods are more likely to fall into overfitting of the trainset and inconsistent with the actual performance, while optimization results of the label-based method on the trainset can better reflect the actual performance, which further denotes the effective and advantageous for the adjustment towards the soft labels.

\section{Experiments}
\label{sec:Experiments}

\subsection{Experimental Settings}
We conduct our experiments on three benchmark datasets: CIFAR-10 \cite{krizhevsky2009learning}, CIFAR-100, and Tiny-ImageNet \cite{le2015tiny}.

\textbf{Baselines.}  We consider the standard training method and eight state-of-the-art methods as comparison methods: \textbf{AT methods}: SAT \cite{madry2017towards}, and TRADES \cite{zhang2019theoretically}; \textbf{ARD methods}: RSLAD \cite{zi2021revisiting}, and AdaAD \cite{huang2023boosting}; \textbf{Robust Fairness methods}: FRL \cite{xu2021robust}, BAT \cite{sun2023improving},  CFA \cite{wei2023cfa}, WAT \cite{li2023wat}, DAFA \cite{lee2024dafa}, and Fair-ARD \cite{Yue2023Revisiting}. It should be mentioned that FRL, CFA, and DAFA include both re-weighting and re-margining operations, while BAT, WAT, and Fair-ARD only include re-weighting operation.

\textbf{Student and Teacher Networks.} For the student model, here we consider two networks for CIFAR-10 and CIFAR-100 including ResNet-18 \cite{he2016deep} and MobileNet-v2 \cite{sandler2018mobilenetv2}. For the teacher model, we follow the setting in \cite{zi2021revisiting,zhao2022enhanced}, and we select WiderResNet-34-10 \cite{zagoruyko2016wide} trained by \cite{zhang2019theoretically} for CIFAR-10 and WiderResNet-70-16 trained by \cite{gowal2020uncovering} for CIFAR-100. 

\textbf{Training Setting.} 
As highlighted in \cite{wei2023cfa}, the worst class robust accuracy fluctuates significantly with the average robustness convergence. Therefore, we adhere to the approach in \cite{wei2023cfa} and choose the best checkpoint based on the highest mean value of all-class average robustness and the worst class robustness (defined as the worst class for CIFAR-10 and the worst-10\% class for CIFAR-100) across all baselines and our method to ensure a fair comparison.

For the baselines, we strictly adhere to the original settings if without additional instruction. For FRL \cite{xu2021robust}, we implement the Reweight+Remargin method with a threshold of 0.05. For CFA \cite{wei2023cfa}, we select the best variant (TRADES+CFA) as indicated in the original paper. During the training of CFA on CIFAR-100, we set the fairness threshold at 0.02 for the FAWA operation, based on the robustness of the worst-10\% class. For Fair-ARD \cite{Yue2023Revisiting}, we choose the best versions (Fair-ARD for CIFAR-10 and Fair-RSLAD for CIFAR-100) as reported in the original paper. For the natural model, we train for 100 epochs, and the learning rate is divided by 10 at the 75-th and 90-th epochs.

\begin{table*}[t] \scriptsize
\caption{Result in average robustness(\%) (Avg.$\uparrow$), worst-10\% robustness(\%) (Worst$\uparrow$), and normalized standard deviation (NSD$\downarrow$) on CIFAR-100 of ResNet-18.}
\centering  
\scalebox{1.05}  { 
\begin{tabular}{c|ccc|ccc|ccc|ccc|ccc}
\hline
\multirow{2}{*}{Method} & \multicolumn{3}{c|}{Clean}     & \multicolumn{3}{c|}{FGSM}      & \multicolumn{3}{c|}{PGD} & \multicolumn{3}{c|}{CW$_{\scriptscriptstyle \rm{\infty}}$}      & \multicolumn{3}{c}{AA}         \\ \cline{2-16} 
                        & Avg.  & Worst & NSD        & Avg.  & Worst & NSD        & Avg.   & Worst & NSD        & Avg.  & Worst & NSD        & Avg.  & Worst & NSD        \\ \hline
Natural                 & 75.17 & 53.80 & 0.155 & 7.95  & 0     & 1.106           & 0      & 0     & -              & 0     & 0     & -             & 0     & 0     & -              \\
SAT\cite{madry2017towards}                     & 57.18 & 29.30 & 0.292          & 28.95 & 5.60  & 0.607          & 24.56  & 3.20  & 0.682          & 23.78 & 3.10  & 0.697          & 21.78 & 2.30  & 0.747          \\
TRADES\cite{zhang2019theoretically}                     & 55.33 & 28.40 & 0.303          & 30.50 & 7.30  & 0.559           & 27.71  & 5.60  & 0.655          & 24.33 & 3.50  &   0.736        & 23.55 & 3.20  & 0.756          \\
RSLAD\cite{zi2021revisiting}                  & 57.88 & 29.40 & 0.302          & 34.50 & 9.20  & 0.550          & 31.19  & 7.40  & 0.598          & 28.13 & 4.70  &    0.669       & 26.46 & 3.70  & 0.703          \\
AdaAD\cite{huang2023boosting}                   & 58.17 & 29.20 & 0.301          & 34.29 & 8.50  & 0.566          & 30.49  & 6.60  & 0.623          & \textbf{28.31} & 5.00  &  0.683         & \textbf{26.60} & 4.20  & 0.721         \\ \hdashline
FRL\cite{xu2021robust}                     & 55.49 & 30.30 & 0.282          & 28.00 & 7.20  & 0.559           & 24.04  & 5.00  & 0.642           & 22.93 & 3.70  &   0.673       & 21.10 & 2.80  & 0.721          \\
BAT\cite{sun2023improving}                     & 62.71 & 34.40 & 0.267          & 33.41 & 7.90  & 0.555          & 28.39  & 5.30  & 0.626          & 23.91 & 3.00  &    0.716       & 22.56 & 2.50  & 0.755          \\
CFA\cite{wei2023cfa}                      & 59.29 & 33.80 & 0.245          & 33.88 & 10.30 & 0.520          & 31.15  & 8.40  & 0.563          & 26.85 & 5.20  &   0.655        & 25.83 & 4.90  & 0.679          \\ 
WAT\cite{li2023wat} & 46.01 & 25.60 & 0.287 & 23.54 & 6.90 & 0.519 & 21.22 & 5.30 & 0.567 & 18.23 & 4.10 & 0.655 & 17.52 & 3.80 & 0.674  \\
DAFA\cite{lee2024dafa}                     & 58.16 & 33.00 & 0.258          & 32.97 & 8.80 & 0.528          & 29.80  & 7.10 & 0.580          & 26.07 & 4.10 & 0.662          & 25.20 & 4.00 & 0.685          \\
Fair-ARD\cite{Yue2023Revisiting} & 57.19 & 29.40 & 0.303 & 34.06 & 8.40 & 0.558 & 30.57 & 7.00 & 0.598 & 27.96 & 4.50 & 0.666 &  26.15 & 3.90 & 0.713 \\
\textbf{ABSLD}           & 56.76 & 31.90 & 0.258          & \textbf{34.93} & \textbf{12.40} & \textbf{0.470} & \textbf{32.44}  & \textbf{10.50} & \textbf{0.500} & 26.99 & \textbf{6.40}  &\textbf{0.578} & 25.39 &\textbf{5.60}  & \textbf{0.601} \\ \hline
\end{tabular}
} \label{tab:3}
\end{table*}

\begin{table*}[t] \scriptsize
\caption{Result in average robustness(\%) (Avg.$\uparrow$), worst-10\% robustness(\%) (Worst$\uparrow$), and normalized standard deviation (NSD$\downarrow$) on CIFAR-100 of MobileNet-v2.}
\centering  
\scalebox{1.05}  { 
\begin{tabular}{c|ccc|ccc|ccc|ccc|ccc}
\hline
\multirow{2}{*}{Method} & \multicolumn{3}{c|}{Clean}     & \multicolumn{3}{c|}{FGSM}      & \multicolumn{3}{c|}{PGD} & \multicolumn{3}{c|}{CW$_{\scriptscriptstyle \rm{\infty}}$}      & \multicolumn{3}{c}{AA}         \\ \cline{2-16} 
                        & Avg.  & Worst & NSD        & Avg.  & Worst & NSD        & Avg.   & Worst & NSD        & Avg.  & Worst & NSD        & Avg.  & Worst & NSD        \\ \hline
Natural                 & 74.86 & 53.90 & 0.150 & 5.95  & 0     & 1.423          & 0      & 0     & -              & 0     & 0     & -              & 0     & 0     & -              \\
SAT\cite{madry2017towards}                     & 56.70 & 22.90 & 0.338          & 32.10 & 7.00  & 0.580          & 28.61  & 5.30  & 0.650          & 26.55 & 3.50  & 0.706          & 24.36 & 2.40  & 0.764          \\
TRADES\cite{zhang2019theoretically}                  & 57.10 & 30.00 & 0.284          & 31.70 & 8.20  & 0.584          & 29.43  & 6.20  & 0.618          & 25.25 & 3.70  & 0.716          & 24.39 & 3.20  & 0.739          \\
RSLAD\cite{zi2021revisiting}                   & 58.71 & 27.10 & 0.311          & \textbf{34.30} & 8.30  & 0.565          & 30.58  & 6.30  & 0.621          & \textbf{28.11} & 4.30  & 0.683          & \textbf{26.32} & 3.30  & 0.724          \\
AdaAD\cite{huang2023boosting}                   & 54.59 & 24.50 & 0.343          & 31.40 & 6.10  & 0.616          & 27.96  & 4.80  & 0.677          & 25.72 & 2.10  & 0.752          & 23.80 & 1.60  & 0.807          \\ \hdashline
FRL\cite{xu2021robust}                     & 56.30 & 26.50  & 0.311          & 31.03 & 8.00  & 0.548          & 27.52  & 6.10  & 0.602          & 25.40 & 3.70  & 0.657          & 23.28 & 2.70  & 0.714          \\
BAT\cite{sun2023improving}                     & 65.39 & 38.70  & 0.228          & 33.31 & 5.30  & 0.577          & 27.08  & 2.90  & 0.683         & 22.80 & 1.50  & 0.790          & 21.28 & 1.10  & 0.823          \\
CFA\cite{wei2023cfa}                     & 59.15 & 34.10  & 0.255          & 32.50 & 8.80  & 0.551          & 29.38  & 7.00  & 0.601          & 25.16 & 4.00  & 0.690          & 23.78 & 3.20  & 0.727          \\
WAT\cite{li2023wat}  & 48.83 & 30.50 & 0.241 & 24.49 & 8.70 & 0.489 & 22.45 & 7.50 & 0.532 & 18.76 & 3.80 & 0.637 & 17.96 & 3.50 & 0.664 \\
DAFA\cite{lee2024dafa}                     & 53.80 & 27.60 & 0.293          & 29.48 & 6.70 & 0.572          & 27.08  & 5.10 & 0.615          & 22.62 & 2.90 & 0.730          & 21.78 & 2.30 & 0.755          \\
Fair-ARD\cite{Yue2023Revisiting} & 58.97 & 31.40 & 0.289 & 33.76 & 9.30 & 0.542 & 30.07 & 7.30 & 0.607 & 27.64 & 5.60 & 0.651 & 25.79 & 4.00 & 0.700 \\
\textbf{ABSLD}           & 56.66 & 32.00  & 0.252          & 33.87 & \textbf{12.80} & \textbf{0.448} & \textbf{31.24}  & \textbf{11.40} & \textbf{0.489} & 26.41 & \textbf{7.50}  & \textbf{0.564} & 24.57 & \textbf{6.70}  & \textbf{0.597} \\ \hline
\end{tabular}
} \label{tab:4}
\end{table*}

For ABSLD, the model is trained using the Stochastic Gradient Descent (SGD) optimizer, starting with an initial learning rate of 0.1, a momentum value of 0.9, and a weight decay of 2e-4 .  We apply 300 epochs for training on both CIFAR-10 and CIFAR-100. The learning rate is reduced by a factor of 10 at the 215th, 260th, and 285th epochs, we set the batch size to 128 for both datasets following \cite{zi2021revisiting}. Additionally, the learning rate $\eta_{\tau}$ for temperature begins at 0.1, respectively, the teacher temperatures $\tau_{k}^{t}$ and $\Tilde{\tau}_{k}^{t}$ are initially configured to 1 for all classes, while the student temperature $\tau^{s}$ remains fixed at 1 without further adjustment. When additional instructions are provided, the maximum and minimum temperature values are set to 5 and 0.5, respectively. For ResNet-18 on CIFAR-100, these values are adjusted to a maximum of 3 and a minimum of 0.8. In the case of inner maximization, we implement a 10-step PGD with a random starting size of 0.001 and a step size of 2/255, ensuring that the perturbation remains bounded by the $L_{\infty}$ norm with $\epsilon = 8/255$. In addition, as for the sample-based and label-based combination version, $\eta_{w}$ and $\eta_{\epsilon}$ is initially set to 0.05 and 0.2, $\beta$ is set to 2, and $\gamma$ is set to 0.04. The maximum and minimum weights are set to 2.5 and 5/6, The maximum and minimum margin are set to 2$\epsilon$ and 0.98$\epsilon$.


\textbf{Metrics.} 
We use two metrics to assess robust fairness: Normalized Standard Deviation (NSD) \cite{Yue2023Revisiting} and worst class robustness \cite{wei2023cfa}. NSD, introduced in \cite{Yue2023Revisiting}, is calculated as NSD = SD/Avg., where SD represents the standard deviation of class-wise robustness and Avg. denotes the average robustness.
NSD evaluates robust fairness while accounting for average robustness. A lower standard deviation indicates better fairness, while a higher average reflects better robustness, meaning that a smaller NSD indicates superior overall performance in terms of fairness and robustness. Worst class robustness is straightforward: a higher value signifies better fairness. For CIFAR-10, we directly report the worst class robust accuracy. For CIFAR-100 and Tiny-ImageNet, due to the low performance of worst class robustness and the limited number of test images (100 images per class for CIFAR-100 and 50 images per class for Tiny-ImageNet), we follow the approach used in CFA \cite{wei2023cfa} and report the worst-10\% class robust accuracy. Additionally, we provide the average robustness as a reference.

Following previous studies \cite{zi2021revisiting, zhao2023mitigating, Yue2023Revisiting}, we assess the trained model against white-box adversarial attacks: FGSM \cite{goodfellow2014explaining}, PGD \cite{madry2017towards}, and CW${\scriptscriptstyle \rm{\infty}}$ \cite{carlini2017towards}. For PGD, we use 20 steps with a step size of 2/255, and for CW${\scriptscriptstyle \rm{\infty}}$, we use 30 steps with a step size of 2/255. Additionally, we employ a strong attack, AutoAttack (AA) \cite{croce2020reliable}, to evaluate robustness. AutoAttack comprises four components: Auto-PGD (APGD), the Difference of Logits Ratio (DLR) attack, FAB-Attack \cite{croce2020minimally}, and the black-box Square Attack \cite{andriushchenko2020square}. The maximum perturbation for all adversarial examples generated is set to 8/255.

\subsection{Ablation Study}
To certify the effectiveness of our ABSLD, we conduct a set of ablation studies, The experiments are conducted on CIFAR-10 of ResNet-18. 

\setlength{\tabcolsep}{1pt}
\begin{figure}[t]
  \centering
  \includegraphics[width=0.85\linewidth]{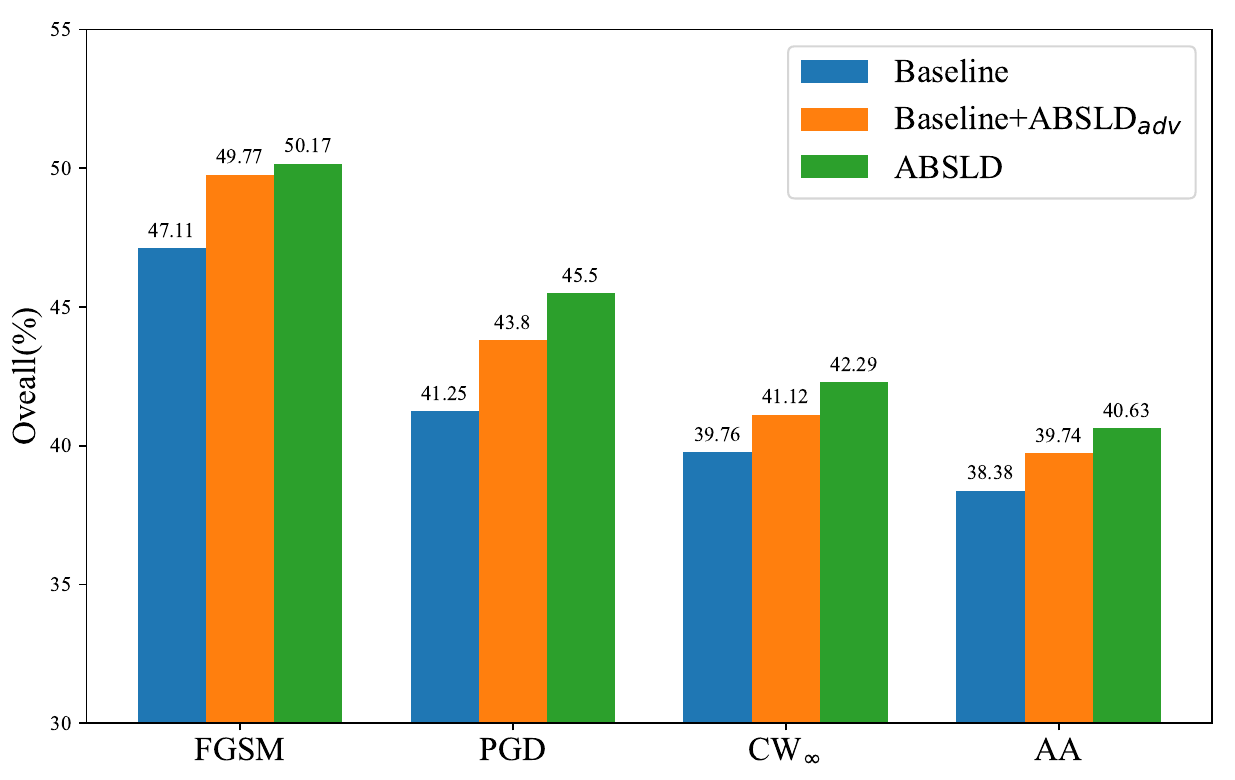}\\
\caption{Ablation study with ResNet-18 student trained using Baseline, Baseline+ABSLD$_{adv}$, and ABSLD on CIFAR-10 of ResNet-18, respectively. }
\label{ablation study}
\end{figure}

\subsubsection{Effects of different components}

We conduct ablation experiments on each component of ABSLD. First, based on the baseline \cite{zi2021revisiting}, we re-temperate the teacher's soft labels for the adversarial examples while keeping the soft labels for clean examples unchanged (Baseline+ABSLD$_{adv}$). Next, we re-temperate the teacher's soft labels for both adversarial and clean examples (ABSLD). The results are illustrated in Figure \ref{ablation study}.  The results demonstrate the effectiveness of ABSLD, and pursuing fairness for clean examples can help robust fairness as claimed in \cite{xu2021robust}.
\begin{figure}[t]
  \centering
\subfloat[\centering Standard deviation of class-wise clean optimization error risk.]{\label{figure_ca3}\includegraphics[width=0.48\linewidth]{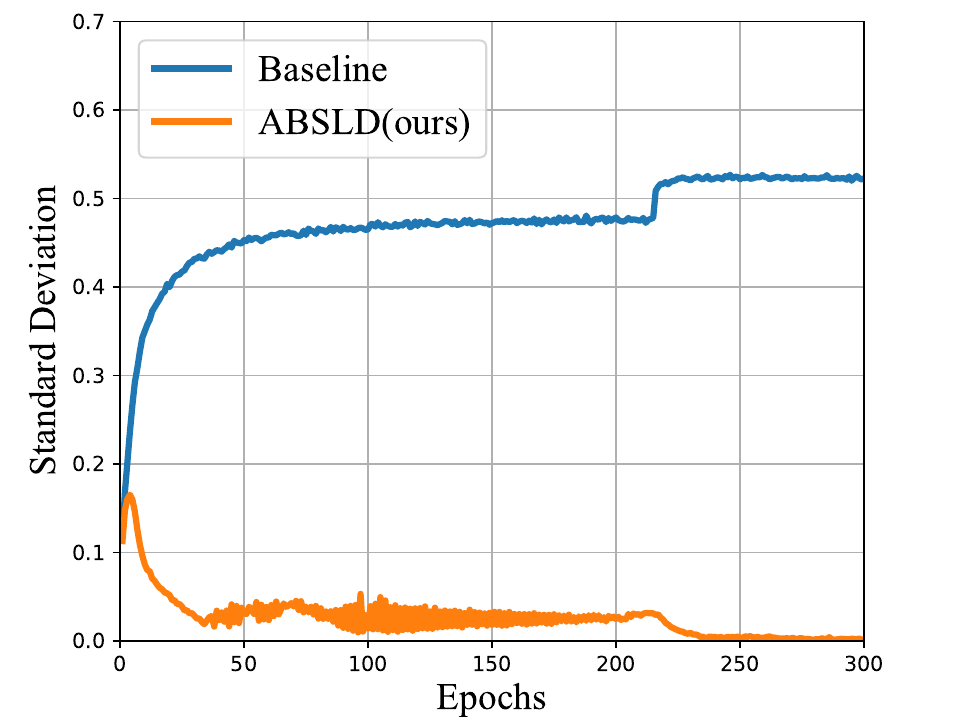}}
\subfloat[\centering Standard deviation of class-wise adversarial optimization error risk.]{\label{figure_cb3}\includegraphics[width=0.48\linewidth]{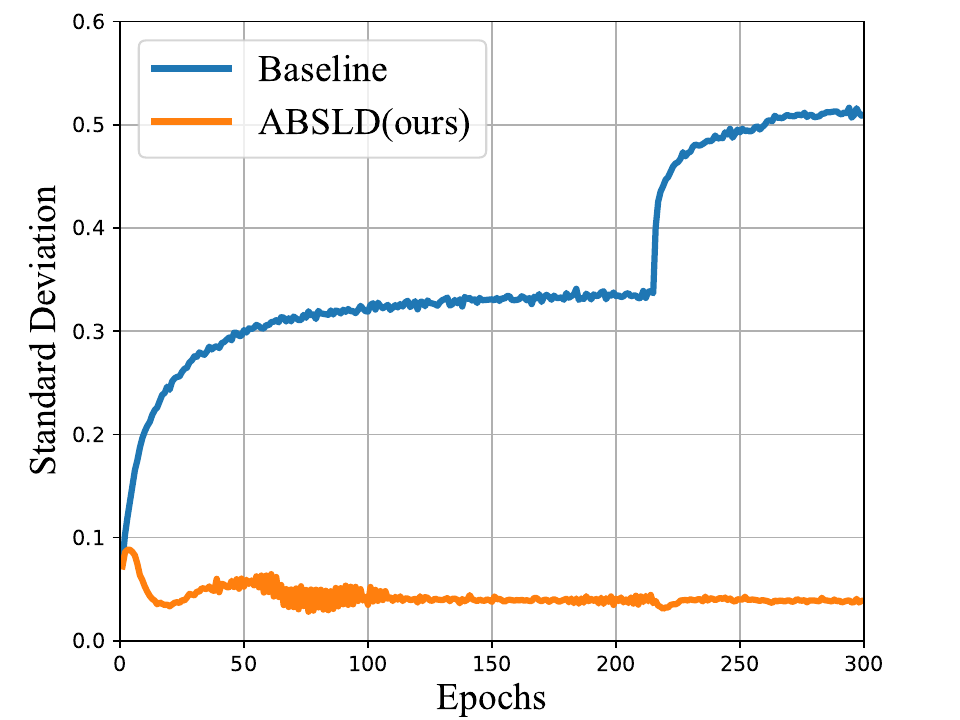}}
\caption{The class standard deviation of student optimization error risk in the training process. 
The result demonstrates that our method remarkably reduces the class-wise optimization error risk gap compared with the baseline during the training process.}
\label{ablation:variance}
\end{figure}

To evaluate the optimization effectiveness of our method, we visualize the standard deviation of class-wise optimization error risk throughout the training process (the optimization error risk is normalized by dividing by the mean), which reflects the optimization gap among different classes. We present the standard deviation for both clean and adversarial optimization error risks, with the results displayed in Figure \ref{figure_ca3} and Figure \ref{figure_cb3}. It is evident that the standard deviation of the baseline increases as the training epochs progress, indicating that the baseline focuses more on reducing the error risk for easier classes while neglecting the harder classes, ultimately resulting in robust unfairness. In contrast, our ABSLD significantly reduces the standard deviation of the student's class-wise optimization error risk, showcasing the effectiveness of our approach.

\subsubsection{Effects of Adaptive Adjustment}
To demonstrate the effectiveness of our self-adaptive temperature adjustment strategy, we 
 manually re-temperate with the static temperature for different classes based on the prior knowledge. Specifically, We set the teacher temperature for difficult class to be small ($\Tilde{\tau}_{k}^t=0.5$) and set the teacher temperature for easy class to be large ($\Tilde{\tau}_{k}^t=5$), which is the minimum and maximum values of temperature, respectively. The experiment in Table \ref{table:manual} shows that our adaptive strategy has a better performance than this manual strategy on CIFAR-10. Moreover, the manual strategy needs to be carefully designed and lacks operability for more complex datasets, e.g., 100 classes on CIFAR-100 or 200 classes on Tiny-ImageNet, so we finally apply the adaptive adjustment strategy for ABSLD.
 
\begin{table}[ht]  \scriptsize
\centering
\tabcolsep=0.155cm
\caption{Result in average robustness(\%) (Avg.$\uparrow$), worst robustness(\%) (Worst$\uparrow$), and normalized standard deviation (NSD$\downarrow$) on CIFAR-10 of ResNet-18.}
\label{table:manual}
\scalebox{1.05}  { 
\begin{tabular}{c|ccc|ccc}
\hline
\multirow{2}{*}{Method}   & \multicolumn{3}{c|}{CW$_{\scriptscriptstyle \rm{\infty}}$}      & \multicolumn{3}{c}{AA}         \\ \cline{2-7} 
                              & Avg.  & Worst & NSD        & Avg.  & Worst & NSD        \\  \hline 
Manual  & 48.13 &29.80  &0.249   & 45.34  & 24.40 & \textbf{0.251}  \\
\textbf{Adaptive}    & \textbf{51.77} & \textbf{32.80} & \textbf{0.249} & \textbf{50.25} & \textbf{31.00} & 0.256 \\ \hline
\end{tabular}
} 
\end{table}

\subsubsection{Effects of Knowledge Distillation}

Here we try to verify the necessity of the application of knowledge distillation (KD). ARD is a type of state-of-the-art adversarial training method currently and can effectively bring strong adversarial robustness for trained models within the framework of KD. Therefore, based on the impressive performance, we can maximally pursue approaches that bring both strong overall robustness and fairness just as in previous work. Meanwhile, Knowledge Distillation itself can bring competitive robust fairness. In Table \ref{tab:1}, Table \ref{tab:2}, Table \ref{tab:3}, and Table \ref{tab:4}, we can notice that KD-based methods themselves (e.g., RSLAD and AdaAD) have competitive robust fairness compared with the other baseline methods (e.g., TRADES, FRL, and CFA). We believe that the KD-based method is more friendly in improving the robust fairness of the model. So we select the ARD as the baseline for better overall robustness and fairness.
\begin{table}[ht] \scriptsize
\tabcolsep=0.155cm
\caption{Comparison between Labelsmoothing, re-temperating via Labelsmoothing (LS), and re-temperating via Knowledge Distillation (KD) in average robustness(\%) (Avg.$\uparrow$), worst robustness(\%) (Worst$\uparrow$), and normalized standard deviation (NSD$\downarrow$).}
\scalebox{1.05}  {
\label{table:LS-KD}
\begin{tabular}{c|ccc|ccc}
\hline
\multirow{2}{*}{Method}  & \multicolumn{3}{c|}{CW$_{\scriptscriptstyle \rm{\infty}}$}      & \multicolumn{3}{c}{AA}  \\ \cline{2-7}
~ & Avg. & Worst & NSD & Avg.  & Worst & NSD\\
\hline
 LabelSmoothing& 48.90 & 20.60 & 0.347 & 47.16& 18.00 &0.368 \\
LS + Retemperating & 47.89 & 27.10  &  0.284 & 46.09 &25.70 & 0.295 \\
\textbf{KD + Retemperating} & \textbf{51.77} & \textbf{32.80} & \textbf{0.249} & \textbf{50.25} & \textbf{31.00} & \textbf{0.256} \\
\hline
\end{tabular}
}
\end{table}

Meanwhile, we also try to explore the performance of using re-temperating directly via label smoothing without a teacher model and add the direct label smoothing operation as the baseline for comparison. The results of Table \ref{table:LS-KD} show that re-temperating directly via label smoothing is better than the baseline, which further demonstrates the effectiveness of re-temperating. Actually, compared with simple label smoothing operations, the teacher's predicted labels contain more information and have a better guiding effect on the student. Therefore, the KD-based method is still superior to the method based on label smoothing.


\subsubsection{Effects of different hyper-parameters}

Here we further discuss the hyper-parameter selection of the initial teacher’s temperature $\tau^{t}$ and the initial temperature learning rate $\eta_{\tau}$ on CIFAR-10 of ResNet-18. The initial teacher temperature setting affects the approximate range of temperature change, while the temperature learning rate $\eta_{\tau}$ influences the temperature change: If the learning rate is too large, the training process will be unstable; If the learning rate is too small, it will not have the effect of regulating the temperature. As Table \ref{table:differnt tau}, and Table \ref{table:different betatau}, the value of $\tau^{t}$ and $\eta_{\tau}$ can slightly influence the final results in a proper range, and the selection of hyper-parameter (initial $\tau^{t}$ as 1, initial $\eta_{\tau}$ as 0.1) is reasonable. 

\begin{table}[ht] \scriptsize
\begin{center}
\tabcolsep=0.15cm
\caption{Discussion about Initial $\tau^{t}$ in average robustness(\%) (Avg.$\uparrow$), worst robustness(\%) (Worst$\uparrow$), and normalized standard deviation (NSD$\downarrow$) on CIFAR-10 of ResNet-18.}
\scalebox{1.05}  {
\begin{tabular}{c|ccc|ccc}
\hline
\multirow{2}{*}{Initial $\tau^{t}$}& \multicolumn{3}{c|}{CW$_{\scriptscriptstyle \rm{\infty}}$}     & \multicolumn{3}{c}{AA} \\ \cline{2-7}
 & Avg.  & Worst & NSD & Avg.  & Worst & NSD  \\ \hline
 0.5   & \textbf{52.21}  & 32.80 & 0.253 & \textbf{50.70} & 30.90 & 0.262 \\
\textbf{1}    &  51.77 & \textbf{32.80} & \textbf{0.249} & 50.25 & \textbf{31.00} & \textbf{0.256} \\
 2    & 51.74 & 32.90 & 0.250 & 50.24 & 30.50 & 0.259  \\
  3    & 48.23 & 31.30 & 0.259 & 47.12 & 29.80 & 0.268 \\ \hline
\end{tabular}
}
\label{table:differnt tau}
\end{center}
\end{table}

\begin{table}[ht] \scriptsize
\begin{center}
\tabcolsep=0.15cm
\caption{Discussion about Initial $\eta_{\tau}$ in average robustness(\%) (Avg.$\uparrow$), worst robustness(\%) (Worst$\uparrow$), and normalized standard deviation (NSD$\downarrow$) on CIFAR-10 of ResNet-18.}
\scalebox{1.05}  {
\begin{tabular}{c|ccc|ccc}
\hline
\multirow{2}{*}{Initial $\eta_{\tau}$}& \multicolumn{3}{c|}{CW$_{\scriptscriptstyle \rm{\infty}}$}     & \multicolumn{3}{c}{AA} \\ \cline{2-7}
 & Avg.  & Worst & NSD & Avg.  & Worst & NSD  \\ \hline
0.05  & 51.53 & 32.50 & 0.252  & 50.11 & 30.40 & 0.256  \\ 
\textbf{0.1} & 51.77 & \textbf{32.80} & \textbf{0.249} & 50.25 & \textbf{31.00} & \textbf{0.256} \\
0.2   & \textbf{51.82} & 32.30 & 0.262  & \textbf{50.42} & 30.10 & 0.269 \\
0.5    & 50.63 & 32.10 & 0.257 & 49.24 & 30.90 & 0.262 \\ \hline
\end{tabular}
}
\label{table:different betatau}
\end{center}
\end{table}

\subsection{Comparisons with SOTA Methods}
 The performances of ResNet-18 and MobileNet-v2 trained by our ABSLD and other baseline methods under the various attacks are shown in Table \ref{tab:1},  Table \ref{tab:2} for CIFAR-10, and Table \ref{tab:3},  Table \ref{tab:4} for CIFAR-100. 
\begin{figure}[t]
\centering
\subfloat[\centering Class-wise Robustness of ResNet-18 on CIFAR-10.]{\label{figure_ca2}\includegraphics[width=0.48\linewidth]{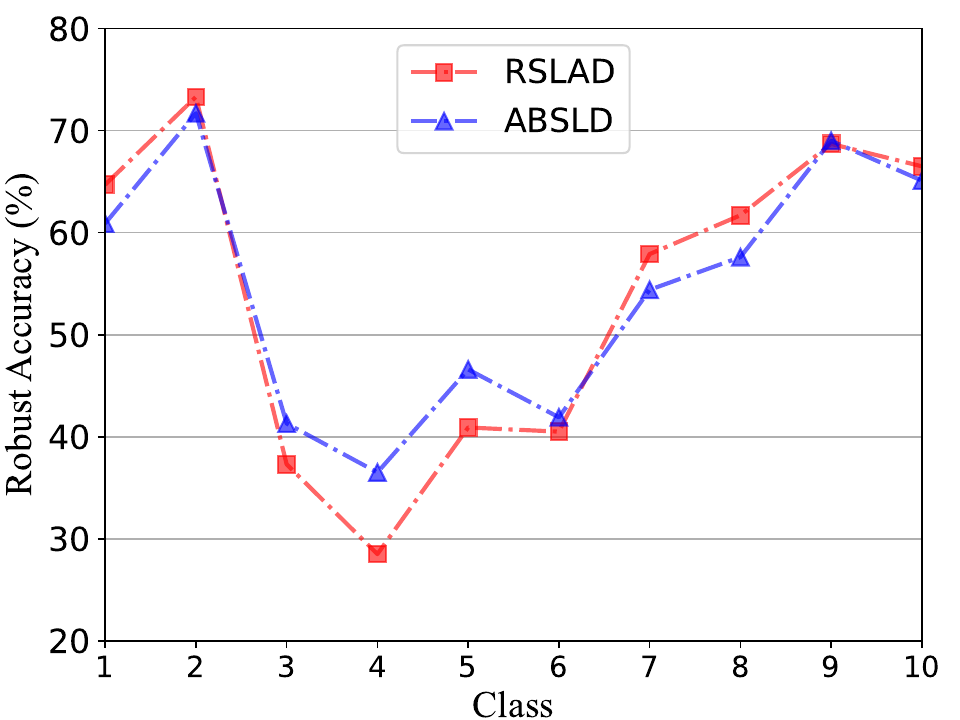}}
\subfloat[\centering Class-wise  Robustness of MobileNet-v2 on CIFAR-10.]{\label{figure_cb2}\includegraphics[width=0.48\linewidth]{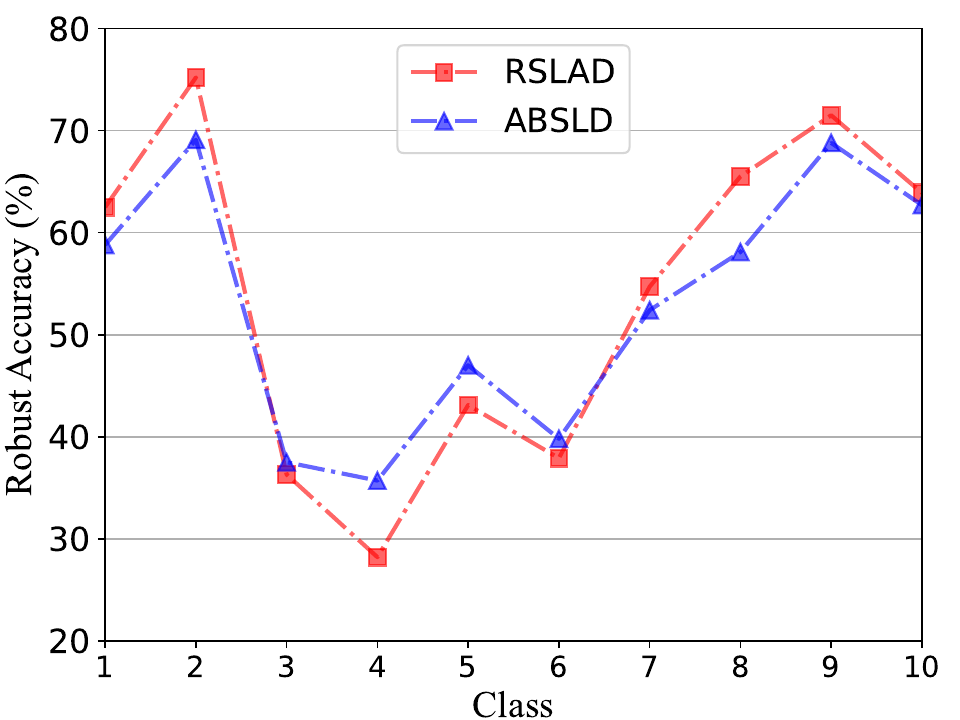}}
\caption{The class-wise robustness (PGD) of models guided by RSLAD and ABSLD on CIFAR-10. It is observed that the robustness of the more challenging classes (class 3, 4, 5, and 6) under ABSLD (blue lines) shows varying degrees of improvement compared to RSLAD (red lines).}
\label{fig:result_class2}
\end{figure}
The results indicate that ABSLD achieves state-of-the-art worst class robustness on CIFAR-10. For ResNet-18 on CIFAR-10, ABSLD enhances the worst class robustness by 0.7\%, 0.7\%, 1.7\%, and 1.5\% compared to the best baseline methods against FGSM, PGD, CW$_{\scriptscriptstyle \rm{\infty}}$, and AA, respectively.
For ResNet-18 on CIFAR-100, ABSLD boosts the worst-10\% class robustness by 2.1\%, 2.1\%, 1.2\%, and 0.7\% compared to the best baseline approach. Additionally, ABSLD demonstrates notable superiority on MobileNet-v2 compared to other methods.


Furthermore, ABSLD achieves the best overall performance in terms of fairness and robustness (NSD) on CIFAR-10. For ResNet-18 on CIFAR-10, ABSLD reduces the NSD by 0.011, 0.016, 0.017, and 0.02 compared to the best baseline method when evaluated against FGSM, PGD, CW$_{\scriptscriptstyle \rm{\infty}}$, and AA, respectively.
On CIFAR-100 with ResNet-18, ABSLD decreases the NSD by 0.049, 0.063, 0.077, and 0.073 compared to the best baseline approach. These results suggest that although the trade-off between robustness and fairness persists, as noted in \cite{ma2022tradeoff}, ABSLD achieves the highest level of robust fairness while incurring the least reduction in average robustness.


We visualize the class-wise robustness in Figure \ref{fig:result_class2}, showing that the robustness of the more challenging classes (class 3, 4, 5, and 6) improves at varying levels. This indicates that our method enhances overall robust fairness but does not just limit the worst class. Additionally, when combined with Figure \ref{fig:result}, it can be observed that the trend of class-wise bias is similar across different training strategies, suggesting that the bias may originate from the dataset itself, further supporting Corollary \ref{corollary1}.

\subsection{Evaluations on Large-scale Dataset}
\label{appendix: Tiny-ImageNet}
We select the subset of ImageNet: Tiny-ImageNet as the additional dataset. We train with PreActResNet-18 with 100 epochs, while other settings are the same as CIFAR-100. For our ABSLD, the teacher model is PreActResNet-34 trained by TRADES \cite{zhang2019theoretically}, and the maximum and minimum values of temperature are 1.1 and 0.9. We select RSLAD \cite{zi2021revisiting} (the baseline method), Fair-ARD \cite{Yue2023Revisiting} (re-weighting method based on the baseline) and CFA\cite{wei2023cfa} (the second-best method proven in Table 1 and Table 3) as the comparison method. The results in Table \ref{table:tinyimagenet} show ABSLD has the best performance under the metric of the worst class robustness and NSD under different attacks, verifying our effectiveness and generalization on large-scale dataset.
\begin{table}[ht] \scriptsize
\centering
\caption{Result in average robustness(\%) (Avg.$\uparrow$), worst robustness(\%) (Worst$\uparrow$), and normalized standard deviation (NSD$\downarrow$) on Tiny-ImageNet of PreActResNet-18.}
\label{table:tinyimagenet}
\scalebox{1}  { 
\begin{tabular}{c|ccc|ccc|ccc|ccc}
\hline
\multirow{2}{*}{Method} & \multicolumn{3}{c|}{Clean}         & \multicolumn{3}{c|}{PGD} & \multicolumn{3}{c|}{CW$_{\scriptscriptstyle \rm{\infty}}$}      & \multicolumn{3}{c}{AA}         \\ \cline{2-13} 
                        & Avg.  & Worst & NSD        & Avg.   & Worst & NSD        & Avg.  & Worst & NSD        & Avg.  & Worst & NSD        \\  \hline 
RSLAD & 47.95 & 15.60 & 0.388 & \textbf{24.68} & 4.10 & 0.668 & \textbf{20.58} & 1.50 & 0.770 &  \textbf{18.78} & 0.90 & 0.830    \\ 
Fair-ARD & 46.58 & 18.80 & 0.376 & 23.81 & 4.30 &  0.676 & 19.97 & 2.70 & 0.782 & 17.73 & 1.20 & 0.888
 \\
CFA &46.75 & 17.50 & 0.398  & 20.73 & 2.30 & 0.770 & 16.83 & 1.20 & 0.896 & 15.99 & 0.90 & 0.936     \\
\hline
\textbf{ABSLD} & 47.70 & 17.90 & 0.342  & 23.41 &\textbf{4.60} &\textbf{0.626} & 19.62 & \textbf{3.40} & \textbf{0.706} &   17.70 & \textbf{2.30} &\textbf{0.772}        \\ \hline
\end{tabular}
} 
\end{table}

\subsection{Evaluation on Common Corruptions}
To further demonstrate that fairness improvement is not limited to adversarial robustness, we have chosen two common corruptions to evaluate comparison methods and ABSLD, including Gaussian Noise and Color Channel Transformation. As in Table \ref{table:common corruption}, ABSLD improves worst class robustness by 0.9\% and 1.8\%, and reduces NSD by 0.012 and 0.01 on CIFAR-10 of ResNet-18 under those two attacks. The results show that ABSLD can effectively transfer adversarial robust fairness towards common corruptions and still maintain pretty performance.

\begin{table}[ht] \scriptsize 
\centering
\tabcolsep=0.15cm
\caption{Results on two common corruptions, for Gaussian Noise(GN) and Colour Channel Transformations(CCT) in average robustness(\%) (Avg.$\uparrow$), worst robustness(\%) (Worst$\uparrow$), and normalized standard deviation (NSD$\downarrow$) on CIFAR-10 of ResNet-18.}
\label{table:common corruption}
\scalebox{1.05}  {
\begin{tabular}{c|ccc|ccc}
\hline
\multirow{2}{*}{Method}& \multicolumn{3}{c|}{GN}     & \multicolumn{3}{c}{CCT} \\ \cline{2-7}
      & Avg.  & Worst & NSD & Avg.  & Worst & NSD   \\ \hline
 RSLAD       & 74.19 & 49.30 & 0.213 & \textbf{69.92} & 53.60 & 0.184 \\
Fair-ARD    & \textbf{75.70} & 53.80 & 0.193 & 69.55 & 52.10 & 0.199 \\
CFA         & 70.90 & 49.80 & 0.207 & 64.74 & 47.90 & 0.172 \\ \hline 
\textbf{ABSLD(ours)} & 74.75 & \textbf{54.70} & \textbf{0.181} & 69.44 & \textbf{55.40} & \textbf{0.162} \\ \hline
\end{tabular}
}
\end{table}

\subsection{Integration with Sample-based Method}
\subsubsection{Comparison with Sample-based Method} Since Fair-ARD and our ABSLD are both KD-based robust fairness methods, we further discuss the difference between Fair-ARD and ABSLD. Fair-ARD is a sample-based method, which applies the least PGD steps for generating adversarial examples as a metric to adaptively re-weight for different classes. Different from adjusting the weights, ABSLD is a label-based method, which applies the optimization error risk as a metric to adaptively re-temperate the label smoothness degree for different classes. 

\begin{table}[ht] \scriptsize
\begin{center}
\tabcolsep=0.155cm
\caption{Comparison between ABSLD and different Fair-ARD's versions in average robustness(\%) (Avg.$\uparrow$), worst robustness(\%) (Worst$\uparrow$), and normalized standard deviation (NSD$\downarrow$) on CIFAR-10 of ResNet-18.}
\scalebox{1.05}  { 
\begin{tabular}{c|ccc|ccc}
\hline
\multirow{2}{*}{Method} & \multicolumn{3}{c|}{CW$_{\scriptscriptstyle \rm{\infty}}$}     & \multicolumn{3}{c}{AA} \\ \cline{2-7}
~ & Avg. & Worst & NSD & Avg.  & Worst & NSD\\
\hline
Fair-ARD& 49.96 & 28.30 & 0.312&  47.97&  25.10 &  0.338 \\
Fair-IAD & 51.22 & 27.50 & 0.300 & 49.58& 25.00& 0.321\\
Fair-MTARD & 51.88 & 24.60 & 0.316 & 50.36& 22.30& 0.330 \\
Fair-RSLAD& \textbf{52.79} &27.40 &0.300 &\textbf{50.69} &25.40& 0.305
 \\
\hline
\textbf{ABSLD(ours)} & 51.77 & \textbf{32.80} & \textbf{0.249} & 50.25 & \textbf{31.00} & \textbf{0.256} \\
\hline
\end{tabular}
\label{Comparison with fair-ard}
}
\end{center}
\end{table}

To further verify the effectiveness, we compare our ABSLD with Fair-ARD's different version on CIFAR-10 of ResNet-18, including Fair-ARD, Fair-IAD, Fair-MTARD, and Fair-RSLAD. Specially, for the comparison between Fair-RSLAD and ABSLD, except for the specific hyper-parameters (e.g., the Fair-RSLAD’s for re-weighting or the ABSLD’s for re-temperating), the baseline (RSLAD) and its corresponding hyper-parameters are completely the same. The results of Table \ref{Comparison with fair-ard} show that ABSLD outperform different types of Fair-ARD versions by 5.6\% and 0.049 in the metric of the worst class robustness and NSD under AutoAttack, which demonstrates the effectiveness and superiority of our ABSLD.

\subsubsection{Combination with Sample-based Method}
We further discuss the combination between the label-based method and the sample-based method. We conduct the experiment on CIFAR-10 of ResNet-18. As the result of Table \ref{table:+reweight}, we find that the combination can achieve better robust fairness no matter with the assistance of sample-based method (Re-weight or Re-margin), and the ABSLD+Re-weight+Re-margin  version achieves the best fairness performance (best worst class robustness and NSD). 

\begin{table}[ht]
\caption{Comparison results about ABSLD, ABSLD+Re-weight version (ABSLD+RW), and  ABSLD+Re-weight+Re-margin version (ABSLD+RW+RM) in average robustness(\%) (Avg.$\uparrow$), worst robustness(\%) (Worst$\uparrow$), and normalized standard deviation (NSD$\downarrow$).}
\tabcolsep=0.145cm
\label{table:+reweight}
\centering
\begin{tabular}{c|ccc|ccc}
\hline
\multirow{2}{*}{Method}   & \multicolumn{3}{c|}{CW$_{\scriptscriptstyle \rm{\infty}}$}      & \multicolumn{3}{c}{AA}         \\ \cline{2-7} 
                          & Avg.   & Worst  & NSD    & Avg.     & Worst    & NSD      \\ \hline
ABSLD     & \textbf{51.77} & 32.80 & 0.249 & \textbf{50.25} & 31.00 & 0.256    \\
ABSLD+RW & 51.55 & 33.20 & 0.235 & 49.96 & 31.80 & 0.241  \\ 
ABSLD+RW+RM & 50.66 & \textbf{36.30} & \textbf{0.227} & 49.17 & \textbf{35.00} & \textbf{0.232} \\
\hline
\end{tabular}
\end{table}

The results demonstrate that the label-based method and sample-based method will mutually promote the improvement of robust fairness and do not conflict with each other. However, we also notice that these methods can improve adversarial fairness, and correspondingly, a slight decrease exists in adversarial robustness, indicating that the trade-off between overall robustness and fairness exists and still needs to be further studied.

\section{Conclusion}
\label{sec:Conclusion}
In this paper, we conducted a thorough investigation into the factors that affect robust fairness during the model optimization process. Initially, we discovered that the smoothness levels of soft labels across various classes can be utilized to mitigate robust fairness issues, supported by empirical observations and theoretical analyses. Subsequently, we introduced Anti-Bias Soft Label Distillation (ABSLD) to tackle the robust fairness challenge by modifying the class-wise smoothness of soft labels. We adjusted the teacher's soft labels by assigning varying temperatures to different classes according to the student's class-wise error risk. A series of experiments demonstrated that ABSLD outperformed state-of-the-art methods in the comprehensive metric (NSD) of robustness and fairness.

\ifCLASSOPTIONcaptionsoff
  \newpage
\fi

\bibliographystyle{splncs04}
\bibliography{egbib}

\vspace{-1cm}
\begin{IEEEbiography}[{\includegraphics[width=1in,height=1.25in,clip,keepaspectratio]{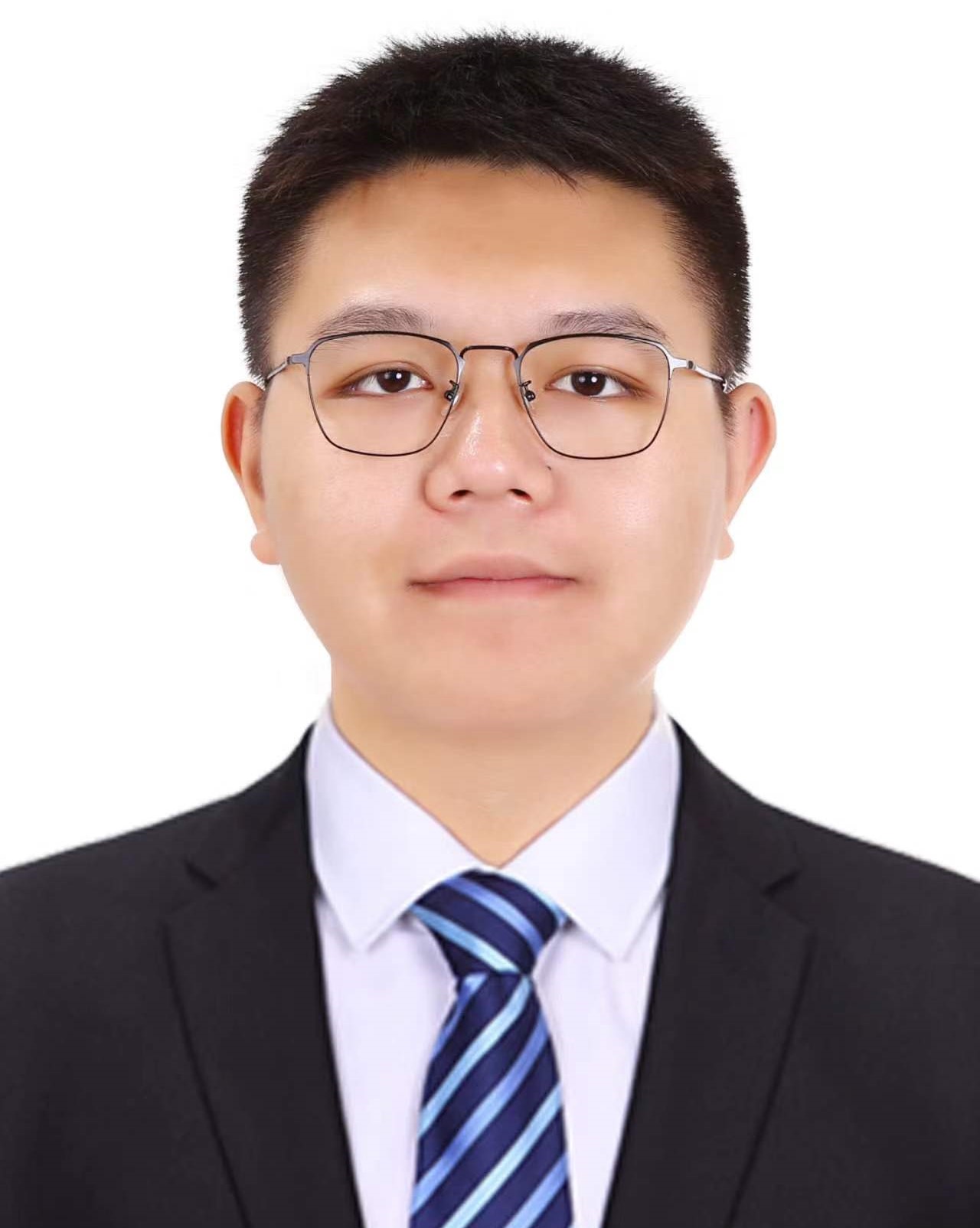}}]{Shiji Zhao} received his B.S. degree in the School of Computer Science and Engineering, Beihang University (BUAA), China. He is now a Ph.D student in the Institute of Artificial Intelligence, Beihang University (BUAA), China. His research interests include computer vision, deep learning and adversarial robustness in machine learning.
\end{IEEEbiography}
\vspace{-1cm}
\begin{IEEEbiography}[{\includegraphics[width=1in,height=1.25in,clip,keepaspectratio]{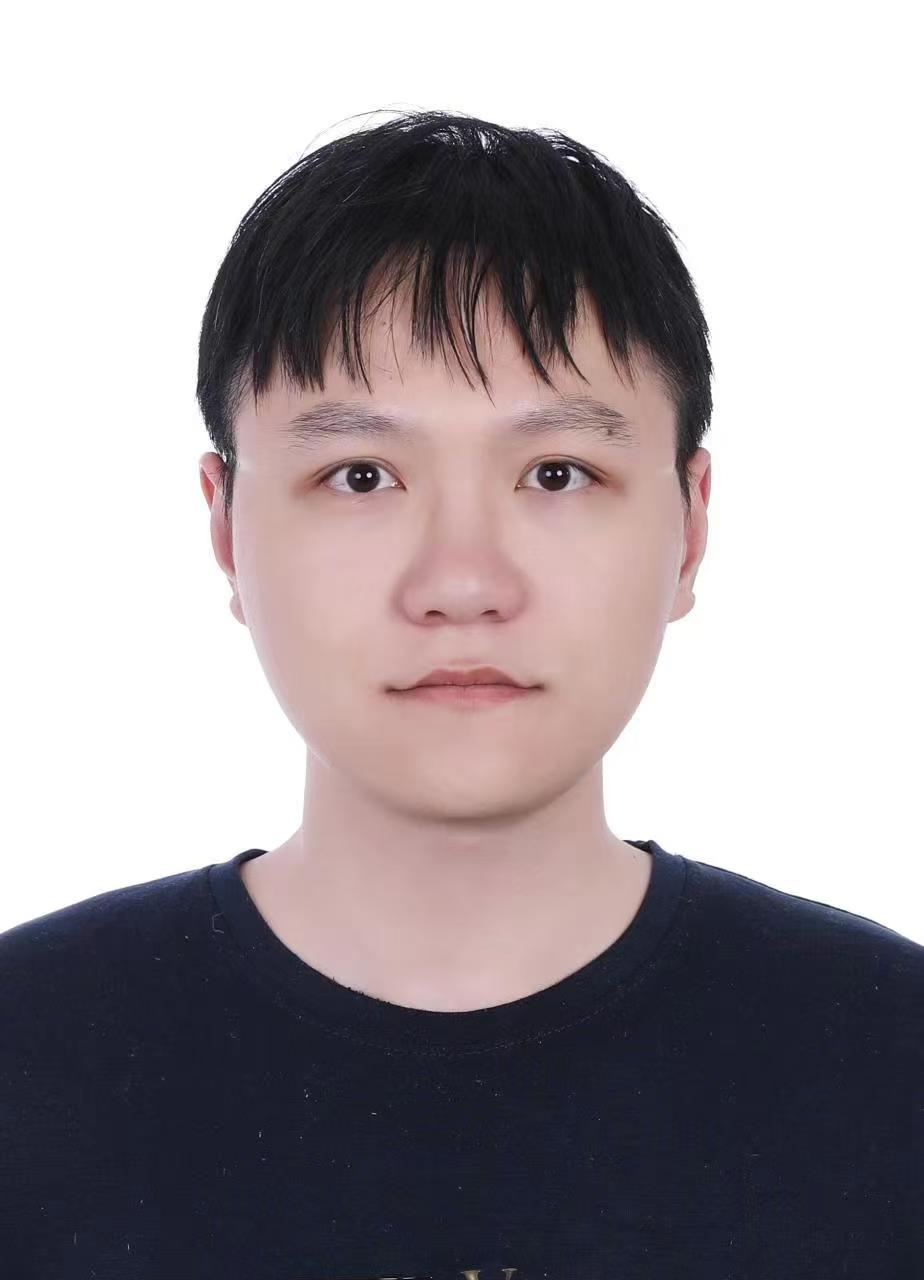}}]{Chi Chen} received his B.S. degree in the School of Computer Science and Engineering, Beihang University (BUAA), China. He is now a M.S. student in the School of Software, Beihang University (BUAA), China. His research interests include computer vision, deep learning and adversarial robustness in machine learning.
\end{IEEEbiography}
\vspace{-1cm}
\begin{IEEEbiography}[{\includegraphics[width=1in,height=1.25in,clip,keepaspectratio]{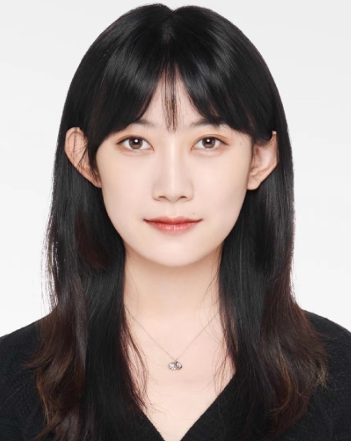}}]{Ranjie Duan} received the B.S. degree from Beijing Institute of Technology and the Ph.D. degree from the Swinburne University of Technology, under the supervision of Prof. Yun Yang. Her long-term goal is to build trustworthy artificial intelligence.
\end{IEEEbiography}
\vspace{-1cm}
\begin{IEEEbiography}[{\includegraphics[width=1in,height=1.25in,clip,keepaspectratio]{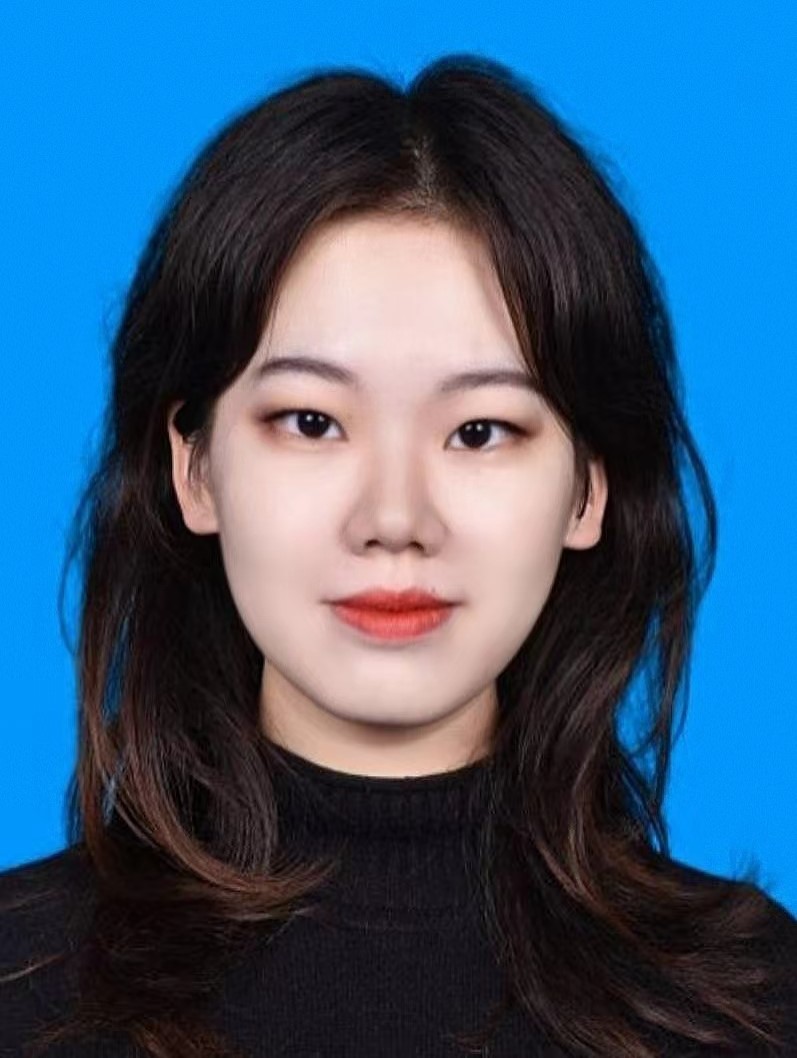}}]{Xizhe Wang} received her B.S. degree in the Institute of Artificial Intelligence, Beihang University (BUAA), China. She is now an M.S. student in the Institute of Artificial Intelligence, Beihang University (BUAA), China. Her research interests include computer vision, deep learning and adversarial robustness in machine learning.
\end{IEEEbiography}
\vspace{-1cm}
\begin{IEEEbiography}[{\includegraphics[width=1in,height=1.25in,clip,keepaspectratio]{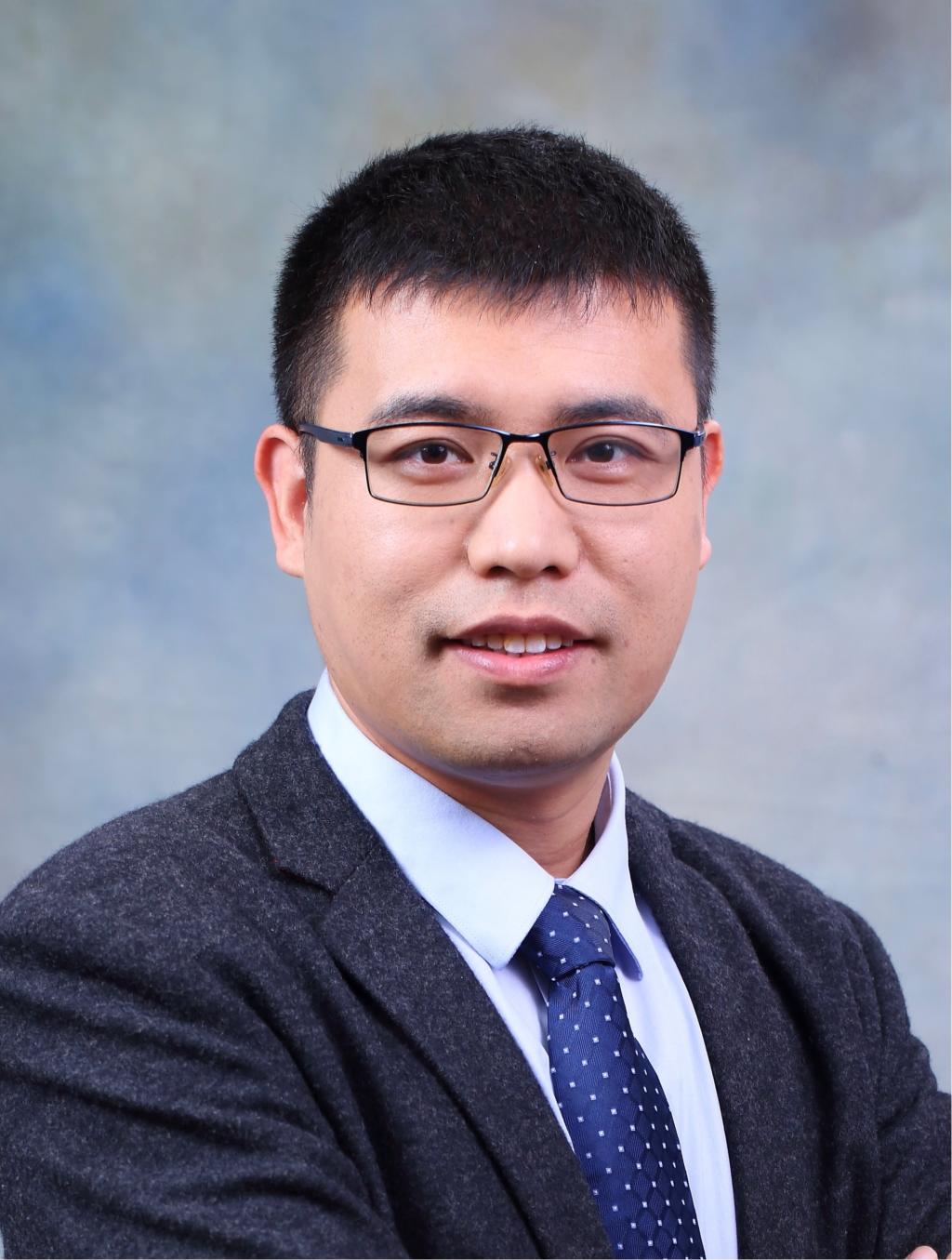}}]{Xingxing Wei} received his Ph.D degree in computer science from Tianjin University, and B.S.degree in Automation from Beihang University (BUAA), China. He is now an Associate Professor at Beihang University (BUAA). His research interests include computer vision, adversarial machine learning and its applications to multimedia content analysis. He is the author of referred
journals and conferences in IEEE TPAMI, TMM, TCYB, TGRS, IJCV, PR, CVIU, CVPR, ICCV, ECCV, ACMMM, AAAI, IJCAI etc.
\end{IEEEbiography}

\end{document}